\documentclass[a4paper,11pt]{article}
\usepackage[utf8]{inputenc}
\usepackage[T1]{fontenc}
\usepackage{graphicx,url}
\usepackage{a4wide}
\usepackage{mathpazo}
\usepackage[font=footnotesize,labelfont=bf]{caption}
\usepackage{amsmath}
\usepackage{amssymb}
\usepackage{multirow}
\usepackage{color,soul}
\usepackage{adjustbox}
\usepackage[shortlabels]{enumitem}
\setlist[enumerate]{nosep}

\usepackage[table,xcdraw]{xcolor}
\usepackage{tabulary}
\usepackage{booktabs}

\title{Using Sequences of Life-events to Predict Human Lives}
\author{Germans Savcisens, Tina Eliassi-Rad, Lars Kai Hansen, Laust
Hvas Mortensen, \\
Lau Lilleholt, Anna Rogers, Ingo Zettler, and Sune Lehmann}

\begin{document}

\maketitle

\begin{abstract}
Over the past decade, machine learning has revolutionized computers' ability to analyze text through flexible computational models \cite{camacho2018word}. 
Due to their structural similarity to written language, transformer-based architectures \cite{rogers2021primer} have also shown promise as tools to make sense of a range of multi-variate sequences from protein-structures \cite{grechishnikova2021transformer, rao2020transformer}, music \cite{huang2018music, zou2022melons}, electronic health records \cite{li2020behrt} to weather-forecasts \cite{chattopadhyay2021towards, bojesomo2021spatiotemporal}. 
We can also represent human lives in a way that shares this structural similarity to language \cite{vafa2022learning}.
From one perspective, lives are simply sequences of events: 
People are born, visit the pediatrician, start school, move to a new location, get married, and so on.
Here, we exploit this similarity to adapt innovations from natural language processing to examine the evolution and predictability of human lives based on detailed event sequences.
We do this by drawing on arguably the most comprehensive registry data in existence, available for an entire nation of more than six million individuals across decades \cite{amrun_data, lpr_data,lynge2011danish, pedersen2011danish}.
Our data include information about life-events related to health, education, occupation, income, address, and working hours, recorded with day-to-day resolution.
We create embeddings of life-events in a single vector space showing that this embedding space is robust and highly structured. 
Our models allow us to predict diverse outcomes ranging from early mortality to personality nuances, outperforming state-of-the-art models by a wide margin. 
Using methods for interpreting deep learning models, we probe the algorithm to understand the factors that enable our predictions.
Our framework allows researchers to identify new potential mechanisms that impact life outcomes and associated possibilities for personalized interventions.
\end{abstract}

\section{Introduction}\label{sec:introduction}
We live in the age of algorithm-driven prediction of human behavior. 
The predictions range from the global and population level, where societies allocate vast resources to predicting phenomena such as global warming \cite{mansfield2020predicting} or the spread of infectious diseases \cite{alali2022proficient}, all the way to the constant flow of individual micro-predictions that shape our reality and behavior as we use social media \cite{zuboff2019age}.
When it comes to individual life outcomes, however, the picture is more complex: 
While it is known that socio-demographic factors play an important role in human lives \cite{weber2009theory}, a collaboration of 160 teams independently analyzing in small groups a comprehensive birth cohort dataset collected over more than 15 years has recently argued that the predictions are typically not accurate, suggesting practical upper limits for predictions of life outcomes \cite{salganik2020measuring}.

Here, we find that with highly detailed data, a different picture of individual-level predictability emerges. 
Drawing on a unique dataset consisting of detailed individual-level day-by-day records  \cite{lynge2011danish, pedersen2011danish}, describing the 6 million inhabitants of Denmark, spanning a 10-year interval, we show that accurate individual predictions are indeed possible.
Our dataset includes a host of indicators, such as health, professional occupation and affiliation, income level, residency, working hours, and education (Methods, Sec.~\ref{sec:data}). 

The central reason we are currently experiencing this age of human prediction is the advent of massive datasets and powerful machine learning algorithms \cite{salganik2019bit, russell2010artificial, grimmer2022text}.
Over the past decade, machine learning has revolutionized image and text processing fields by accessing ever larger datasets that have enabled increasingly complex models \cite{resnet50, hannun2014deep, mnih2013playing}. 
Language processing has evolved particularly rapidly, and transformer architectures have proven successful at capturing complex patterns in massive and unstructured sequences of words \cite{rumelhart1985learning, t5_model, brown2020language}. 
While these models originated in natural language processing, their ability to capture structure in human language generalizes to other sequences \cite{grechishnikova2021transformer, rao2020transformer, huang2018music, zou2022melons, li2020behrt, chattopadhyay2021towards, bojesomo2021spatiotemporal, cai2021msa, choromanski2020rethinking_performer}, which share properties with language, e.g., that sequence ordering is essential, and elements in the sequence can have meaning on many different levels.  
Importantly, due to the absence of large-scale data, transformer models have not been applied to multi-modal socio-economic data outside the industry.

Our dataset changes this. 
The sheer scale of our dataset allows us to construct sequence-level representations of individual human life-trajectories, which detail how each person moves through time.
We can observe how individual lives evolve in the space of diverse types of events (information about a heart attack is mixed with salary increases or information about moving from an urban to a rural area).
The time resolution within each sequence and the total number of sequences are large enough that we can meaningfully apply transformer-based models to make predictions of life outcomes.
This means that representation learning can be applied to an entirely new domain to develop a new understanding of the evolution and predictability of human lives.
Specifically, we adopt a BERT-like architecture \cite{devlin2018bert} to predict two very different aspects of human lives: time of death and personality nuances (additional predictions in SI:~Emigration Tasks).
We find that our model can accurately predict these outcomes, in the case of early mortality, outperforming current state-of-the-art methods by $\sim11\%$, see \textit{Results}.

To make these accurate predictions, our model relies on a single common embedding space for all events in the life-trajectories. 
Just as embedding spaces in language models can be studied to provide a novel understanding of human languages \cite{kozlowski2019geometry, pilehvar2020embeddings}, we can study the concept embedding space to reveal non-trivial interactions between life-events. 
Below, we provide insight into the resulting \textit{concept-space} of life-events and demonstrate the robustness and interpretability of this space and the model itself. 
Transformer-based models also produce an embedding of individuals (the analogy in a language representation is a vector summarizing an entire text). 
Using explainability tools such as saliency maps \cite{ding2019saliency_smooth, atanasova2020diagnostic_saliency} and concept activation vectors (TCAV) \cite{kim2018interpretability}, we show that the person-summaries are also meaningful and hold the potential to serve as a \textit{behavioural phenotype} which can improve other individual-level prediction tasks, for example, to augment analyses of medical images \cite{lucieri2020interpretability}.
Our work has important societal and ethical implications, which we outline in the Discussion as well as in Methods, Sec.~\ref{sec:ethics}, and SI:~Model~Card.

\begin{figure}[ht]
    \centering
    \makebox[\textwidth][c]{\includegraphics[width=1.1\textwidth]{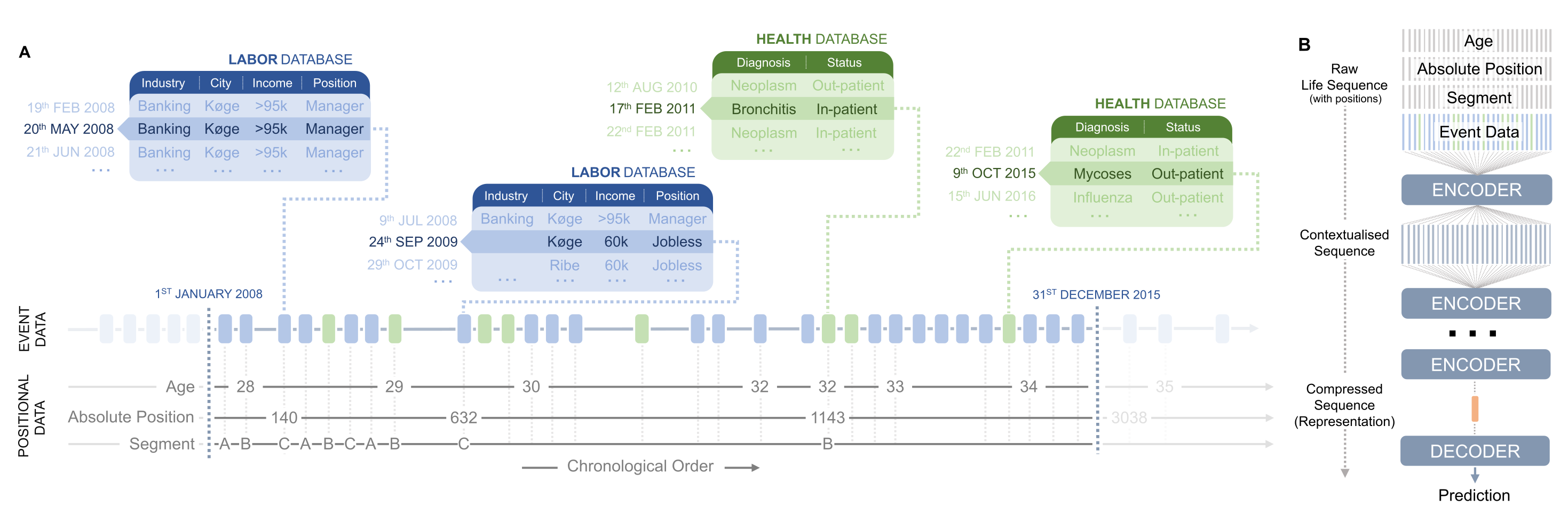}}%
    \caption{\textbf{A schematic individual-level data representation for the \texttt{life2vec} model.} 
    (A) We organize socio-economic and health data from the Danish national registers from 1st January 2008 until 31st December 2015 into a single chronologically ordered \textit{life-sequence}. Each database entry becomes an event in the sequence, where an event has associated positional and contextual data. The contextual data include variables associated with the entry (e.g., ~industry, city, income, job type). The positional data includes the person's age (expressed in full years), absolute position (number of days since 1st January 2008), and segment (alternating sequence of three elements).  The raw life-sequence is then passed to the model described in panel (B). The model consists of multiple stacked encoders. The first encoder combines contextual and positional information to produce a contextual representation of each life event. The following encoders output deep contextual representations of each life event (considering the overall content of the life-sequence). The final encoder layer fuses the representations of life-events to produce the representation of a life-sequence. The decoder uses the latter to make predictions.}
    \label{fig:encoder}
\end{figure}

\section{Results}\label{sec:results}
\subsection{Life-events and Life-sequences}

\textbf{Life-sequences for millions of individuals based on rich data}. 
In the following, we represent the progression of individual lives as \textit{life-sequences} (see Fig.~\ref{fig:encoder}). 
The life-sequences are constructed based on labor and health records from Danish national registers \cite{lynge2011danish, pedersen2011danish}, which contain highly detailed data on work, residence, health, and education for all $\sim$ 6 million Danish citizens. 
Our \textit{labor} dataset \cite{amrun_data} includes records about income, such as salary, scholarship, job-type \cite{ilo_2012}, industry \cite{db07}, social benefits, etc. 
The \textit{health} dataset \cite{lpr_data, lynge2011danish} includes records about initial visits to healthcare professionals or hospitals, accompanied by the diagnosis, patient type, and urgency (encoded according to the ICD-10 system \cite{world1992icd}, SI: Specification of features and their sources). 
Life-sequence evolve over time and provide rich information about life-events with high temporal resolution.

\textbf{We use a simple symbolic language to encode the rich data}.
The raw stream of complex multi-source temporal data poses significant methodological challenges, such as irregular sampling rates, sparsity of data, complex interactions between features, and a high number of dimensions \cite{yadav2018mining}. 
Classical methods for time series analysis (e.g., support vector machines, ARIMA) \cite{han2019review, moncada2021explainable} become cumbersome because they are challenging to scale, inflexible, and require a considerable amount of data preprocessing to extract useful features. 
Using transformer methods allows us to avoid hand-crafted features and instead encode the data in a way that exploits the similarity to language \cite{moncada2021explainable}.
Specifically, in our case, each category of discrete features and discretized continuous features form a \textit{vocabulary}.
This vocabulary -- along with an encoding of time -- allows us to represent each life-event (including its detailed qualifying information) as a \textit{sentence} comprised of synthetic words, or \textit{concept tokens}. 
We attach two temporal indicators to every event. 
One that specifies the individual's age at the time of the event and one that captures absolute time, see Fig.~\ref{fig:encoder}.

Thus, our synthetic language can capture information along the lines of ``In September 2020, Francisco received twenty thousand Danish kroner as a guard at a castle in Elsinore'' or ``During her third year at secondary boarding school, Hermione followed five elective classes''. 
In this sense, the progression of a person's life is represented as a string of such sentences that together form individual \textit{life-sequences}. 
Our approach allows us to encode a wide range of detailed information about events in individual lives without sacrificing the content and structure of the raw data. 

\subsection{The \texttt{life2vec} model}
\textbf{We use transformer models to form compact representations of individual lives.}
We call our deep learning model \texttt{life2vec}.
The \texttt{life2vec} model is based on a transformer-architecture \cite{devlin2018bert, choromanski2020rethinking_performer, bachlechner2020rezero, swish, nguyen2019transformers_scale_norm, pappas2018beyond_weight_tying, weight_tying, smith2019super, loshchilov2017decoupled, liu2019variance}.
Transformers are well suited for representing life-sequences due to their ability to compress contextual information \cite{vaswani2017attention, liu2021end} and take into account temporal and positional information \cite{huang2018music, kazemi2019time2vec}. 

The training of the \texttt{life2vec} consists of two stages.
We first train the model by simultaneously using (1) a Masked Language Modeling (MLM) task that forces the model to use token representations and contextual information \cite{devlin2018bert} and (2) a Sequence Ordering Prediction (SOP) task that focuses on the temporal coherence of the sequence \cite{lan2019albert} (Methods, Sec.:~\ref{sec:train_procedure}). 
The pre-training creates a concept space and teaches the model patterns in the structure of life-sequences, which we discuss below.

Next, to create compact representations of individual life-sequences, the model performs a classification task (Methods, Sec.:~\ref{sec:train_procedure}).
The person-summaries the model learns in this last step is \textit{conditional} on the classification task; it identifies and compresses patterns that maximize the certainty around a given downstream task \cite{jawahar2019does}. 
For example, when we ask the model to predict a person's personality nuances, the person embedding space will be structured around key dimensions that contribute to personality.

\subsection{Accurate predictions across diverse domains}
\label{subsec:predictions}
The first critical test of any model is predictive performance.  
Here, \texttt{life2vec} outperforms the state-of-the-art while simultaneously being able to perform classification in very different domains.
We test our framework on two distinct tasks.

\textbf{Predicting early mortality}.
We estimate the likelihood of a person surviving the following four years after 1st January 2016. This is an oft-used task within statistical modeling~\cite{naemi2021machine}. 
Further, mortality prediction is closely related to other health-prediction tasks and therefore requires \texttt{life2vec} to model the progression of individual health-sequences as well as labor history to predict the right outcome successfully.
Specifically, given a sequence representation, \texttt{life2vec} infers the likelihood of a person surviving the four years following the end of our sequences (1st January 2016). 
We focus on making predictions for a young cohort of people consisting of individuals who are 30-55 years old, where mortality is challenging to predict. 

This prediction task has an additional level of complexity as data contains people with unknown outcomes (i.e., emigrants and missing individuals). 
We account for this issue by applying positive-unlabeled learning \cite{jiang2020improving, wang2021asymmetric}, which gives us a robust loss function for training, as well as a corrected performance metric for the model evaluation. 

\begin{figure}[h]
    \centering
    \makebox[\textwidth][c]{\includegraphics[width=1.0\textwidth]{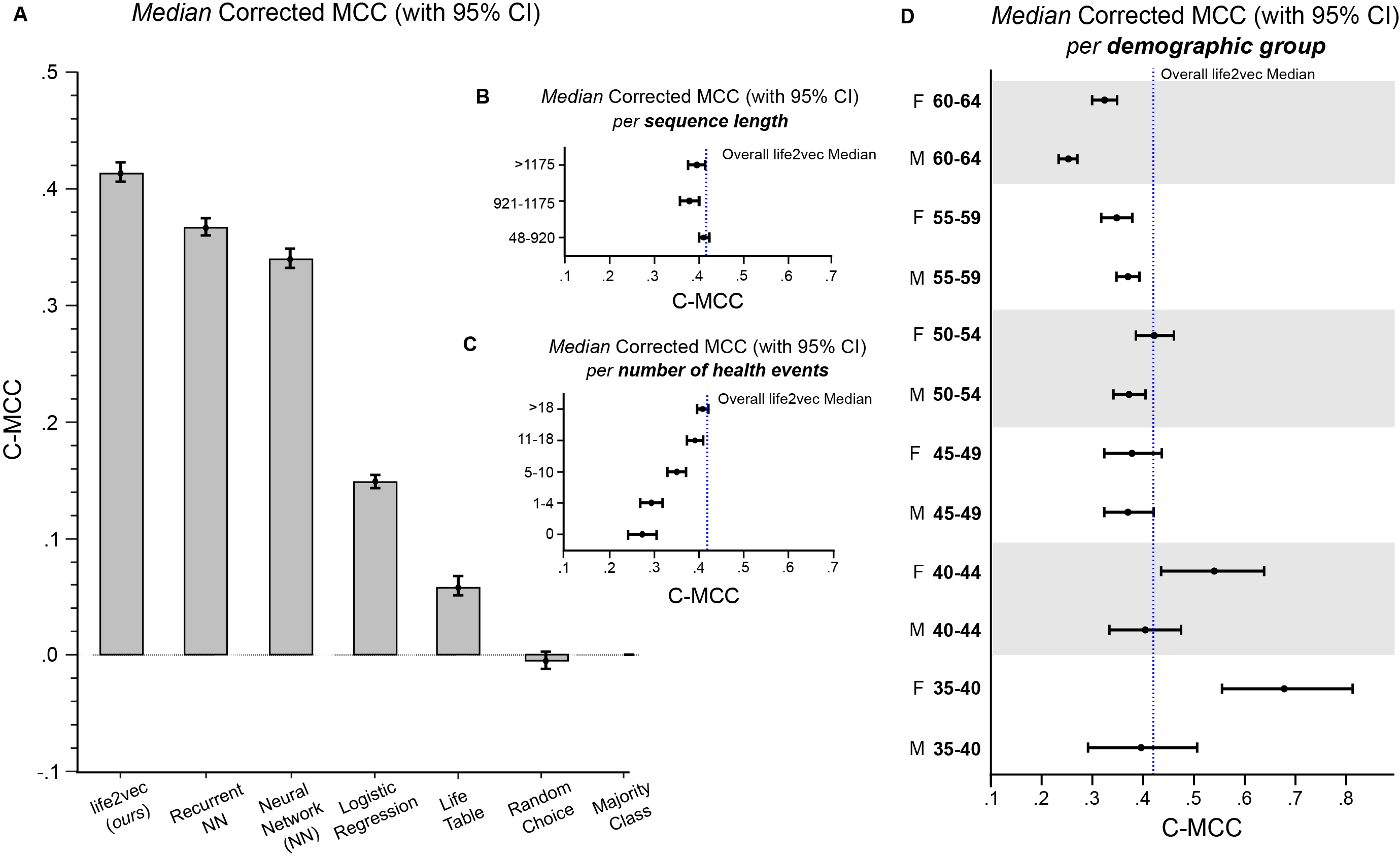}}
    \caption{\textbf{Performance of models on the Mortality Prediction Task quantified with the Median Corrected Matthews correlation coefficient (C-MCC) \cite{ramola2018estimating} with 95\% CI.} (\textbf{A}) Comparison of \texttt{life2vec} performance to baselines (\textbf{B-D}) Performance of \texttt{life2vec} model on different cohorts of the population. (\textbf{B}) Performance of \texttt{life2vec} per sequence length. We can see that sequence length does not affect the performance. (\textbf{C}) Performance of \texttt{life2vec} based on the number of health events in a sequence. The model performs better on cohorts with a higher number of health events. (\textbf{D})  Performance of \texttt{life2vec} per inter-sectional groups (based on age group and sex).}
    \label{fig:performance}
\end{figure}

The performance of \texttt{life2vec} in relation to a range of baseline models \cite{hansen2023predicting}—actuarial life tables, logistic regression, feed-forward neural networks, and recurrent neural networks, is shown in Fig.~\ref{fig:performance} and summarized in Tab.~\ref{tab:eos_metric}.

We illustrate the performance of models using the Corrected Matthews correlation coefficient, C-MCC \cite{chicco2020advantages, ramola2018estimating} (Methods, Sec.:~\ref{sec:methods_evaluation}) that adjusts the MCC value due to the presence of unlabeled samples.
With the median C-MCC Score of 0.41 (95\%~CI [0.40, 0.42]), \texttt{life2vec} outperforms the baselines by 11\% (see Fig.~\ref{fig:performance}); note that increasing the size of RNN models does not improve their performance.
Fig.~\ref{fig:performance}.D also breaks down performance for various sub-groups: intersectional groups based on age and sex, as well as groups based on the sequence length (SI:~Model~Card). 

In terms of age and gender, the model performs better on a younger cohort of the population and on a cohort of females. 
Further, sequence length (i.e., a proxy for a number of life-events in a sequence) does not have a significant impact on the performance of a model (Fig.\,\ref{fig:performance}B). 

\textbf{Predicting personality nuances}.
Death as a prediction target is well-defined and eminently measurable.
To test the versatility of \texttt{life2vec}, we now predict \textit{personality nuances}, an outcome at the other end of the measurement spectrum, something which is internal to an individual and typically measurable through questionnaires.
In spite of the difficulty in measurement, personality is an important feature that shapes people's thoughts, feelings, and behavior and predicts life outcomes \cite{roberts2007power}. 
Specifically, we focus on personality nuances in the domain of the Introversion-Extraversion dimension (for simplicity, Extraversion in what follows) because the corresponding personality nuances are part of virtually all comprehensive models of the basic personality structure that have emerged (in the Western world) over the last century, including the Big Five \cite{mccrae2008five} and HEXACO \cite{zettler2020nomological} frameworks, but also Eysenck's \cite{eysenck1978superfactors} and Jung's \cite{jung1923types} personality models. 
We align the prediction of personality nuances by \texttt{life2vec} with recent research that highlights the advantages of personality nuances (i.e., responses to specific personality questionnaire items) over broader summarizing (i.e., responses across items) personality `facets' (e.g., Extraversion-Social Self-esteem) and `domains' (e.g., Extraversion) in terms of associations with life outcomes \cite{mottus2022leveraging, seeboth2018successful, stewart2022finer}. 
As our dataset, we draw on data collected for a large and largely representative group of individuals in `The Danish Personality and Social Behavior Panel' (POSAP) study \cite{posap} (see Methods Sec.~\ref{sec:data}).
We randomly pick one item (personality nuance) per Extraversion facet and predict individual-level answers. 
 
\begin{figure}[h]
    \centering
    \includegraphics[width=0.6\columnwidth]{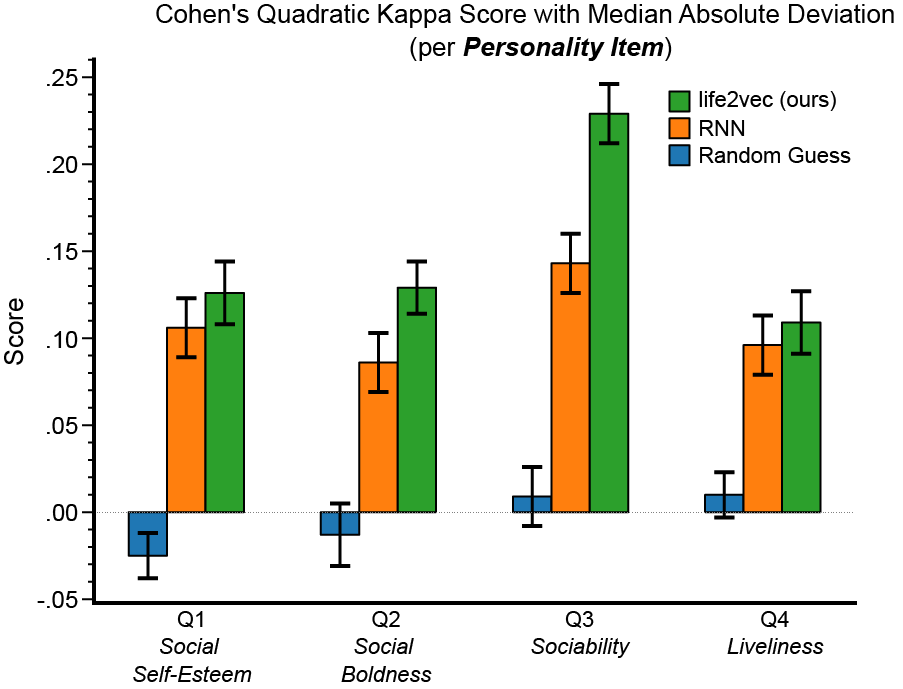}
    \caption{\textbf{Performance Evaluation for the Personality Nuances Task}. We display Cohen's Quadratic Kappa score for each item separately for Random Guess, RNN, and \texttt{life2vec} model. The error bars indicate the Median Absolute Deviation. The question wordings are as follows. Q1 (Social Self-esteem): ``I feel reasonably satisfied with myself overall''. Q2 (Social Boldness): ``When I'm in a group of people, I'm often the one who speaks on behalf of the group''. Q3 (Sociability): ``I prefer jobs that involve active social interaction to those that involve working alone'' Q4 (Liveliness): ``On most days, I feel cheerful and optimistic''.}
    \label{fig:personality_scores}
\end{figure}

Fig.~\ref{fig:personality_scores} shows that applying \texttt{life2vec} to life-sequences not only allows us to predict early mortality but is versatile enough also to capture personality nuances (see Methods Sec.~\ref{met:finetuning}). 
\texttt{life2vec} has better scores than RNN on all items, but the difference is only statistically significant on Items 2 and 3 (see Fig.~\ref{fig:personality_scores} for item wording).
The fact that an RNN trained for this specific task is also able to extract a signal around personality underscores that -- while transformer models are powerful -- a large part of what makes \texttt{life2vec} so versatile is the dataset itself.

We have illustrated \texttt{life2vec}'s versatility with further prediction tasks (SI:~Emigration Task).

\subsection{Concept Space: Understanding relations between concepts}
The building blocks of \texttt{life2vec} are the concept tokens of our synthetic language. 
A key novelty of our approach is that the algorithm learns a single joint multidimensional space that contains all events that can occur in human life. 
We start our exploration of this space with a visualization.

\begin{figure}[h]
    \centering
    \makebox[\textwidth][c]{\includegraphics[width=1.25\textwidth]{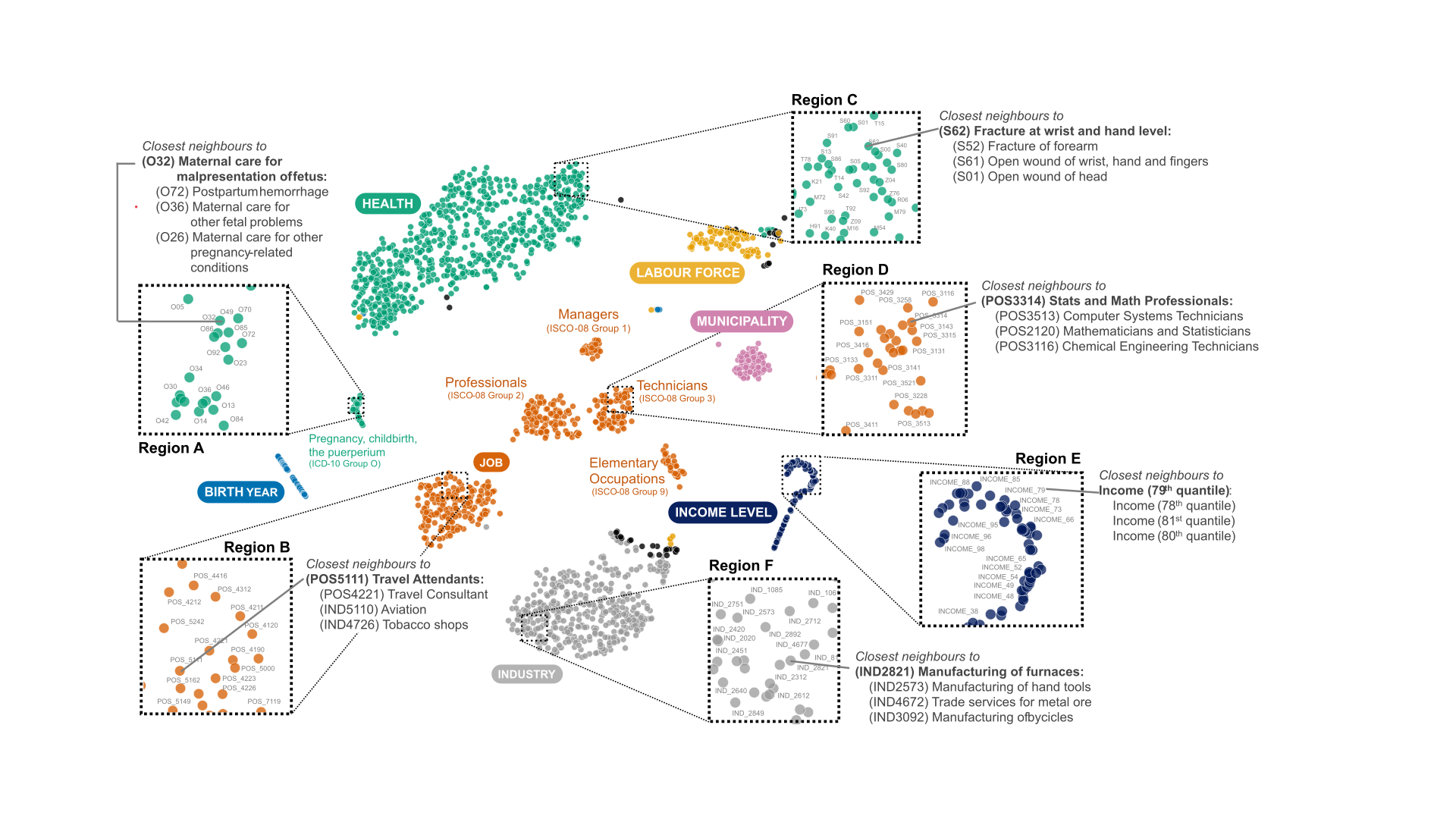}}
    \caption{Two-dimensional projection of the concept space (using the PaCMAP \cite{wang2021understanding}). Each point corresponds to a concept token in the vocabulary. Points are colored based on the concept types (several types are omitted - black points). Each region provides a closer look at several parts of the concept space. You can also see the top three closest neighbors for selected tokens (based on the cosine distance). (\textbf{A}) Diagnoses related to Pregnancy, childbirth, and the puerperium in ICD-10 \cite{world1992icd}. (\textbf{B}) Job concepts related to Service and Sales Workers (corresponds to Job Category 5 of ISCO-08 \cite{ilo_2012}). (\textbf{C}) Injury-related diagnoses in ICD-10 \cite{world1992icd}. (\textbf{D}) Job concepts related to Technicians and Associate Professionals (corresponds to Job Category 3 of ISCO-08 \cite{ilo_2012}). (\textbf{E}) Income-related concepts. \texttt{life2vec} arranges these concepts in increasing ordinal order. (\textbf{F}) Concepts related to the manufacturing industry in DB07 \cite{db07}. 
    }
    \label{fig:embeddings}
\end{figure}

\textbf{The global view}.
In Fig.~\ref{fig:embeddings}, the original 280-dimensional concepts are projected onto a two-dimensional manifold with the use of PaCMAP \cite{wang2021understanding}, that preserves the local and global structures of the high-dimension space. PaCMAP constructs the graph consisting of three types of edges -- that connect neighbors, mid-near pairs, and further pairs. 
These edges define how forces of attraction and repulsion should move points along the two-dimensional manifold~\cite{wang2021understanding}.

Here, each concept is colored according to its type. 
This coloring makes it clear that the overall structure is organized according to the key concepts of the synthetic language: health, job type, municipality, etc., but with interesting sub-divisions, separating a birth year, income, social status, and other key demographic pieces of information. 
The structure of this space is highly robust and emerges reliably under a range of conditions (see Methods Sec.~\ref{subchap:test_concept_space}).

\textbf{The fine structure of concept space is meaningful}.
Digging deeper than the global layout, we find that the model has learned intricate associations between nearby concepts. 
We investigate these local structures via neighbor analysis, which draws on the cosine distance between concepts in the original high-dimensional representations as a similarity measure. 
A key place to consider is the cluster formed by income (dark blue points in Fig.~\ref{fig:embeddings}). 
What the model sees is 100 concept tokens, each describing a level of income -- but before training, it has no \textit{a priori} idea of what each one means.
It is simply an arbitrary string of text among other strings, but from training on the life-sequences, the model not only learns that income is different from other concepts (the dark blue points are isolated), but it also perfectly sorts the 100 levels.
The blue curve starts with the token corresponding to the first percentile salaries and organizes them up to the 100th.
Thus, the concepts most similar to the 59th percentile of income are the 58th and the 60th.
Similarly, for birth years (light blue in Fig.~\ref{fig:embeddings}): the closest concepts to the birth year 1963 are 1962 and 1964, and so on.

The health-type cluster (green points in Fig.~\ref{fig:embeddings}) has a solid local structure. Diagnoses belonging to the same ICD-10 \cite{world1992icd} chapters cluster according to their chapter. For example, the concept `malignant neoplasm of stomach' (C16 in ICD-10) is surrounded by other C-Chapter concepts, such as `malignant neoplasm of lungs' (C34) and `malignant neoplasm of colon' (C18). As shown in Fig.~\ref{fig:embeddings}A, one of the clearly separated health-clusters relates to pregnancies and childbirth diagnoses (i.e., O-Chapter concepts).

The concepts of professional occupation also cluster into smaller groups. These groups roughly correspond to the Major Groups of the International Standard Classification of Occupations (ISCO-08) \cite{ilo_2012}. Clearly defined clusters exist for 1st (Managerial and Executive Positions), 2nd (Professionals), 3rd (Technicians and Associate Professionals), and 9th (Elementary Occupations) groups.

Not all concept tokens are surrounded by tokens of the same category, but even in these cases, the neighborhoods are meaningful. 
In Fig.~\ref{fig:embeddings}B job-concept of a `travel agent' is surrounded by the job-concept of a `travel consultant' and an industry-concept of Aviation.
When the model does mix up ICD-10 codes, the `mistakes' are meaningful. 
For example, the concept of Z95 (Presence of cardiac and vascular implants and grafts) is surrounded by concepts corresponding to ICD-10 Chapter I \cite{world1992icd}, for example, I42 (Cardiomyopathy), I50 (Heart failure), and I25 (Chronic ischemic heart disease). 
The model's ability to group similar concepts that are not necessarily close in the standard classification systems is one of the strengths of our approach. 
Understanding which life-events play equivalent roles in human lives is one of the aspects which allow for improved classification and recommendation.

\subsection{Person-summaries: Understanding the representation of individuals}
Along with the concept representations described above, \texttt{life2vec} creates dense representations of individual life-sequences, \textit{person-summaries}. 
The person-summary is a single vector that encapsulates the essential aspects of an individual's entire sequence of life-events; the person-summaries span our person embedding space. 
To form a person-summary, the model determines which aspects are relevant to the task at hand. 
In this sense, the person-summaries are conditioned on a specific prediction task. 
Below, we focus on person-summaries for the case of mortality likelihood, but person-summaries relative to, e.g., change in the area of residence or choice of the university would be drastically different.

\begin{figure}[h]
    \centering
    \makebox[\textwidth][c]{\includegraphics[width=1.2\textwidth]{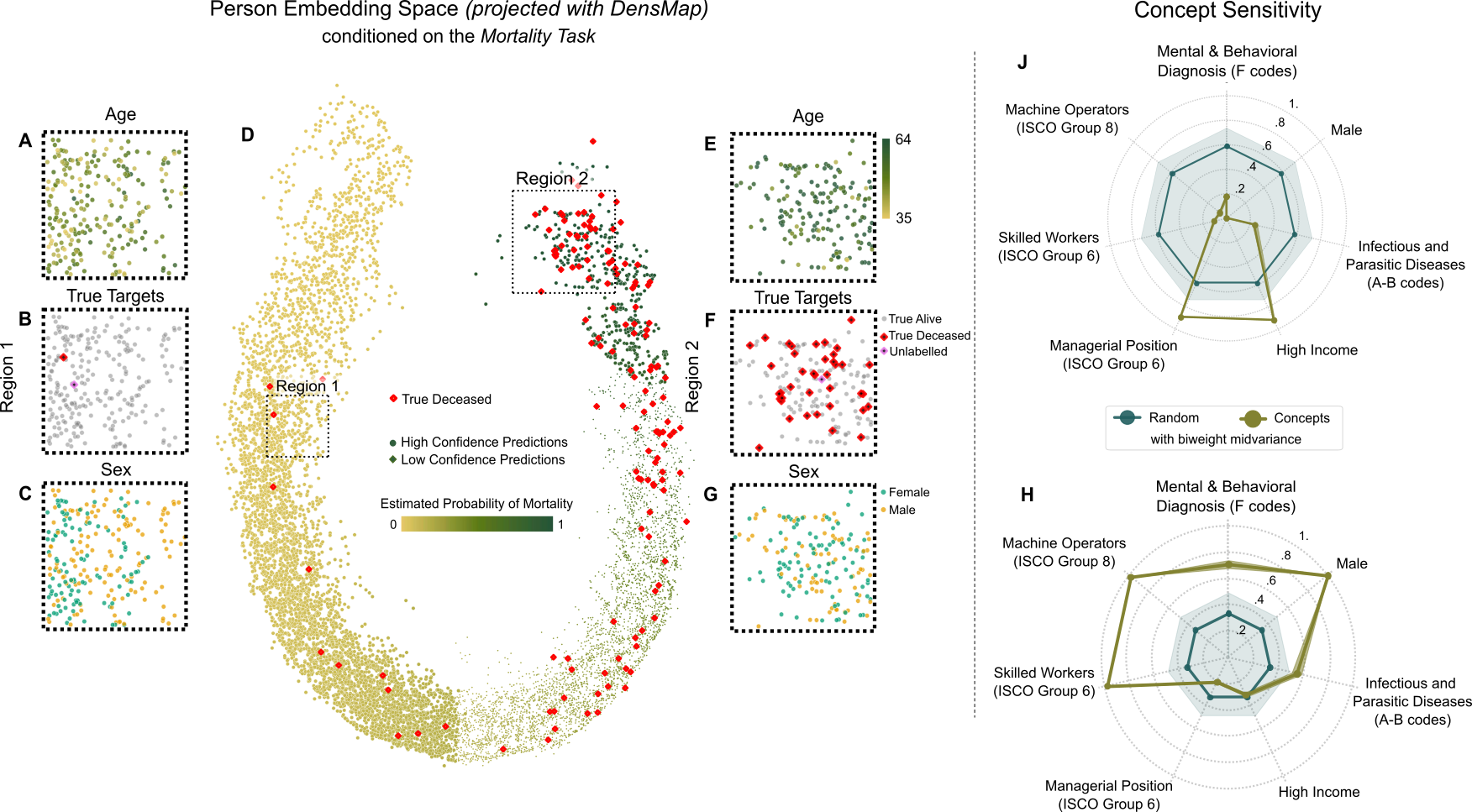}}
    \caption{Representation of life-sequences conditioned on the Mortality Predictions. (\textbf{A-G}) Two-dimensional projection of 280-dimensional life representations(with the DensMap method \cite{narayan2020density}). (\textbf{D}) The full projection is colored based on the estimated probability of mortality. Pink points stand for the true deceased targets. Points with a smaller radius are uncertain predictions. (\textbf{A-C} and \textbf{E-G}) Zoomed-in regions with additional aspects associated with the life-sequence. (\textbf{A-C}) Region A contains points with a low probability of mortality, while (\textbf{E-G}) Region B contains points with a high probability. (\textbf{J-H}) Spider plot of \texttt{life2vec}'s concept sensitivity. The blue line is a median score for the random concept directions, while the blue area specifies the variation of the scores for the random concepts (\textbf{J}) Concept Sensitivity with respect to "Alive" prediction. (\textbf{H}) Concept sensitivity with respect to the "Deceased" prediction. }
    \label{fig:attribution}
\end{figure}

\textbf{Overview of the person-summaries}.
The space of person-summaries is visualized in Fig.~\ref{fig:attribution}\,A-G.
Relative to the mortality prediction, the model organizes individuals on a continuum from low to the high estimated probability of mortality (the point cloud in panel D).
In Fig.~\ref{fig:attribution}, we show true deceased through purple diamonds, while the confidence of predictions \cite{geifman2017selective} is demonstrated via the radius of points (e.g. dots with a small radius are low-confidence predictions). 
Further, the estimated probability is displayed using a color map from yellow to green.  
We zoom in on two regions: Region 1, which shows an area with a high probability of the `survive' outcome, and Region 2, with a high probability of the `death' outcome. 
We see that while Region 2 has a majority of elderly individuals, we still see a large fraction of younger individuals (Fig.~\ref{fig:attribution}\,E) and that it contains a fraction of true targets (Fig.~\ref{fig:attribution}\,F). 
Region B has a largely opposite structure, with a majority of young individuals but a substantial number of older individuals as well (Fig.~\ref{fig:attribution}\,E) and only a single actual death (Fig.~\ref{fig:attribution}\,F).
When we look into actual deaths in the low probability region, we find that the five deaths nearest to and in Region 1 have the following causes -- two accidents, malignant neoplasm of the brain (C71.9), malignant neoplasm of cervix uteri (C53.8), and myocardial infarction (I21.9), all causes of death that we would expect to be difficult to predict from life-event sequences.

\textbf{Directions in the person embedding space using TCAV}.
Topic Concept Activation Vectors (TCAV) \cite{kim2018interpretability}, give us a way to understand the meaning of directions in the person embedding space using labeled data.
The idea behind TCAV is to use binary labeled data (e.g., the labels `employed'/`unemployed') and identify the hyperplane that best separates those labels. 
The vector orthogonal to this hyperplane gives us a direction for `employed'-`unemployed' in the embedding space (the Concept Activation Vector \cite{kim2018interpretability}).
We then use this employment-direction to understand how that label impacts decisions. 
Specifically, we measure how moving our decision boundary along this direction changes predictions; how the prediction reacts to these changes is called the \textit{concept sensitivity}.

Fig.~\ref{fig:attribution}\,J,H show concept sensitivity scores for several labels relative to the mortality-prediction task.
Here we show a two-dimensional projection using DensMap \cite{narayan2020density}, but a range of other low-dimensional projections (T-SNE \cite{van2008visualizing}, UMAP \cite{mcinnes2018umap}, PaCMAP \cite{wang2021understanding}) are available in SI (Sec.: Visualisation of Embedding Spaces).
We focus on health-related labels such as a history of mental disease (or its absence), nervous system disease, diagnosis of neoplasm, and `endocrine, nutritional and metabolic diseases.' 
Similarly, we use socio-economic attributes as labels -- to measure the model's sensitivity to major occupational groups, sex, education, and origin. 
Fig.~\ref{fig:attribution}\,J shows labels in relation to the prediction `survive', and Fig.~\ref{fig:attribution}\,H shows concepts with respect to the prediction `death' within the four years following our sequence.
Values close to one imply that moving in the topic direction indicates that moving in the label-direction increases the probability of a specific outcome.  
Values close to zero indicate the opposite. 
The gray areas are what we would expect if we moved in a random direction. 
We see that directions of possessing a managerial position or having a high income nudge the model towards the 'survive' decisions (Fig.~\ref{fig:attribution}\,J), while being male, a skilled worker, or having a mental diagnosis has the opposite effect (Fig.~\ref{fig:attribution}\,H).
Note that while the spider plots in Fig.~\ref{fig:attribution}\,J,H are almost mirrors, they are created based on different data sets, a further validation of robustness.

To confirm the validity of the sensitivity scores, we further perform extensive significance testing (Methods, Sec.~\ref{subchap:methods_interpretability}).
Our final approach to understanding the person-summaries is via inspection of the model's attention to individual sequences \cite{atanasova2020diagnostic_saliency, ding2019saliency_smooth, bastings2020elephant} -- these confirm the findings discussed above (SI:~Interpretability). 

\section{Discussion}
Drawing on the progress from the natural language processing that made ChatGPT \cite{openai2023gpt4}  possible and a massive nation-scale dataset that captures small and large events in the lives of millions of individuals over a decade, the \texttt{life2vec} model builds complex contextual representations of a range of aspects that characterize human lives: health, occupation, geography, and wealth.

When we draw on these representations to make predictions, transformer-based \texttt{life2vec} is able to adapt to different settings, from death-prediction to personality nuances, yielding highly accurate predictions that outperform state-of-the-art baselines trained on the same dataset.

When we investigate how the model can make these predictions, we find that to solve these diverse tasks, the model relies on different aspects of life trajectories. 
Mortality prediction requires the model to estimate how single events impact future outcomes while predicting personality nuances extracts information from large-scale patterns in the trajectories.
More than that,\texttt{life2vec} handles the distinct complications of each task, such as missing labels, imbalanced sample sizes, and ordinal multi-label settings. 

We can shed further light on what the algorithm learns by studying its embedding spaces. 
The highly structured concept embedding space contains the model's fundamental building blocks. 
Here, we show that the model captures a meaningful and robust relationship between tokens of the vocabulary. 
Clusters emerge structured around concept tokens. 
Tokens tend to cluster according to classification systems (e.g., ICD-10, ISCO-08), revealing local relationships (how highly related tokens relate to one another) as well as global (how high-level concept-groups relate to one another) semantic relations in the system.

The model also captures the ordinal nature of features such as time, year, and income. 
Finally, the model converges to a similar embedding space given different subsets of data (and space is not biased with respect to frequent tokens).

In the person embedding space, the model produces representations that condense signals from the entire life-sequence into a single vector.
These representations are always conditioned on specific prediction tasks.
We can probe the person embedding space to gain intuition on why the model makes a certain prediction. 
Here, we find that in many cases model relies on relevant information (health, age, and income for the mortality prediction). 
However, we can also identify less obvious patterns, such as the role of the job-type. 
We can use the insights drawn from these summaries to generate new hypotheses and as a starting point for studies that focus on causality.

In summary, \texttt{life2vec} opens a range of possibilities within the social and health sciences. 
Through a rich dataset, we capture a wealth of complex patterns and trends in individual lives and represent their stories in a compact vector representation.
These vectors represent a new type of comprehensive linkage between social and health outcomes. 
The output of our model, coupled with causality tools, shows a path to (a) systematically explore how different data modalities are correlated and interlinked and (b) use these interlinkages to explicitly explore how life impacts our health and vice versa.
In this sense, we open the door to a new and more profound interplay between the social and health sciences.
Finally, we stress that our work is an exploration of what is possible but should only be used in real-world applications under regulations that protect the rights of individuals (see Methods, Sec.~\ref{sec:ethics}).

\newpage
\begin{small}
\section{Methods}\label{sec:methods}

\subsection{Ethics and Broader Impacts}\label{sec:ethics}
The data analysis was conducted at \textit{Statistics Denmark},  the Danish National Statistical Institution. 
The data analysis was conducted under the Danish Data Protection Act and the General Data Protection Regulation (GDPR)~\cite{gdpr2016}. 
In this context, since the data was used for scientific/statistical purposes, the usage is partially exempt from the GDPR~\cite{gdpr2016} (e.g.~from the right to be forgotten). 
Danish-based academic researchers, government agencies, NGOs, and private companies can be given access to Statistics Denmark data, but access is only granted under strict information security and data confidentiality policies\footnote{\url{https://www.dst.dk/en/OmDS/strategi-og-kvalitet/datasikkerhed-i-danmarks-statistik}} that ensure that data on individual entities are not leaked or used for purposes other than scientific/statistical. 
This focus on safekeeping data is shared with most other National Statistical Institutions that provide similar services. 
Using scientific/statistical `products' such as \texttt{life2vec} for automated individual decision-making, profiling, or accessing individual-level data that may be memorized by the model is strictly disallowed. 
Aggregate statistics, including those coming from model predictions, may be used for research and to inform policy development.

We stress that \texttt{life2vec} is a research prototype, and in its current state, it is not meant to be deployed in any concrete real-world tasks. 
Before it could be used, e.g., to inform public policies in Denmark, it should be audited, in particular, to ensure the demographic fairness \cite{10.1145/3457607} of its predictions (with respect to the appropriate fairness metrics for the given context) and explainability \cite{burkart2021survey} (e.g.~if used for assisting decision-making based on synthetic/counterfactual data). 
Such audits would likely soon be mandated by the AI Act\footnote{\url{https://www.europarl.europa.eu/thinktank/en/document/EPRS_BRI(2021)698792}}, focusing on the safe use of 'high-risk' models. 
Further auditing information is located in SI:~Model~Card.

Finally, we note that while it is possible that phenomena captured by \texttt{life2vec} reflect phenomena that have similar distributions outside of Denmark (e.g., labor market trajectories, individual health trajectories) --  we urge caution with extrapolation to other populations since we have not explored how our findings translate beyond the current study population.

\subsection{Dataset}\label{sec:data}
We work with the Labour Market Account (\texttt{AMRUN}) \cite{amrun_data} and the National Patient Registry (\texttt{LPR}) datasets \cite{lynge2011danish, world1992icd}. Within the Labour Market Account dataset are event data for every resident of Denmark. For Danish residents who have been in contact with secondary of health care services, primarily hospitals, the events are accounted in the National Patient Registry.  We limit ourselves to data recorded in the period from 2008 until the end of 2015. Datasets are pseudonymized prior to our work by de-identifying addresses, Central Person Register numbers (CPRs), and names. Data is stored within Statistics Denmark, and all access/use of data is logged.

The total number of residents in the filtered dataset is \textit{3\,252\,086}.  For our research, we choose people who (1) are alive and lived in Denmark on the 31st December 2015, (2) have at least 12 records in the labor data during the year of 2015\footnote{Corresponds to 12 incomes over one year (e.g.~salary, pension, etc.). We do not set requirements on the health-set as not every resident has any records in the health dataset}, (3) have consistent sex and birthday attributes over the whole residency period, (4) are between 25 and 65 years old on the 31st December 2015.

These prerequisites apply for both stages: pre-training and finetuning (mortality prediction and self-reported personality questionnaires).

For the mortality prediction task, we excluded young individuals with very low death rates and older individuals with a high background probability of death. Thus, we narrowed the specification of requirements (4) and limited the dataset to people who are between 35-55 years old on 31st December 2015 (which limits us to \textit{2\,301\,993} people).

For the personality nuances prediction task, we do not alter the initial requirement (4) but add new requirements on top of the original ones: (5) residents should have participated in the POSAP study~\cite{posap}, and (6) none of the scores associated with any HEXACO personality nuance (facet, dimension) are missing. This results in analyzing responses of \textit{9\,794} people. 

Specifically, in POSAP HEXACO-60 \cite{ashton2009short} was administered, comprising 60 items (each representing one personality nuance) that can further be aggregated in (24) personality facets and, in turn, six personality dimensions (Honesty-Humility, Emotionality, Extraversion, Agreeableness vs. Anger, Conscientiousness, Openness to Experience). 
T
\subsubsection{Labour Data}
The Labour Marked Accounts dataset \cite{amrun_data} contains data on each taxable income a resident receives, such as a salary, state scholarship, pension, etc. Each taxable income has multiple associated features, we focus on 16 features, see Tab.~\ref{tab:variables}. Some of these features are linked to the workplace: \textit{Type of Enterprise} \cite{eurostat2013european}, \textit{Industry Code} \cite{db07}. Others describe personal attributes: \textit{Professional Positions} \cite{ilo_2012}, \textit{Labour Force Status}, \textit{Labour Force Status Modifier}, \textit{Residential Municipality}, \textit{Income}, \textit{Working hours}, \textit{Tax Bracket}, \textit{Age}, \textit{Country of Origin} and \textit{Sex}.

Types of Enterprise feature is based on \textit{European system of accounts (ESA2010)} \cite{eurostat2013european}, while Industry codes are encoded with Danish Industry Code (DB07) \cite{db07}. Industry codes provide information about the type of services the company offers. For example, code 108400 stands for the `Preparation of flavorings and spices', and 643040 stands for the `Venture companies and private equity funds'. ESA2010 has an intrinsic structure, which allows us to use more general categories (i.e., only the first four digits of a code).

Job types are classified via the \textit{International Standard Classification of Occupations} (ISCO-08) \cite{ilo_2012}. The system encodes job types with four digits, e.g., code 2111 references `physicists and astronomer', while code 5141  references `barbers'. However, several codes exceed the length of 4, and since ISCO-08 also has hierarchies, we can collapse those to four-digit codes.

Labour Force Status provides information about a person's attachment to the Labour Market. The attachment does not solely include different forms of employment. For example, for a person enrolled in an official higher educational program, the status would be a `student'. Being unemployed is also a type of attachment, even though the financial compensation is not a salary. Some labor force statuses have additional information in the form of a modifier. If present, the modifier gives specifications for the labor force status. If the labor force status is student, the modifier might specify a `foreign student'. A person can have multiple labor force statuses in the same period of time. Using the student example again, a student can also have employment alongside studying, and both would be accounted for in the dataset.

Since we want to have a concept token representation of continuous variables, such as income and labor-force-period, we binarize them based on quantiles. For example, the income variable is split into 100 categories. Another continuous variable is the labor-force-period. It is a percentage of days in a month that the Labour Force status is relevant for (binned in 10 categories). We also reserve concept tokens for each birth year and birth month.

\subsubsection{Health Data}
The health data pertains to all ambulatory and inpatient contacts with hospitals in Denmark. The country has a publicly funded healthcare system that caters to all citizens. The data is encoded using the ICD-10 System \cite{world1992icd}, an internationally authorized WHO system for classifying procedures and diseases. This system encompasses approximately 70,000 procedures and 69,000 diseases, each term represented by up to 7 symbols. The first symbol denotes the chapter, which represents a specific type of diagnosis. The first three symbols combined provide the category. For example, code \texttt{S86} is in chapter \texttt{S}, which stands for the `injuries and poisoning' and \texttt{S86} combined stands for the `injury of muscle, fascia, and tendon at lower leg level'. By adding or removing symbols, one can control the specificity of the term.

To reduce the vocabulary size, we collapsed all codes to the category level, which resulted in 704 terms. The data includes patient type, emergency status, and urgency in addition to diagnoses. Patient type denotes the admission type, i.e., inpatient, outpatient, or emergency. Emergency status indicates a patient admitted via an Emergency Care Unit, while urgency specifies whether the cause of admission was an acute onset.

\subsubsection{Preprocessing}
Each health and labor record is translated into a sentence, where each associated attribute (e.g., diagnosis, job type) is converted to a concept token. For example, if a labor record is connected to a job type `Work with archiving and copying' (code \texttt{9210} in ISCO-08 \cite{ilo_2012}), we convert it to \texttt{POS\_9210}. As a result, we have two types of sentences: \textit{labor sentences} and \textit{health sentences}. For each resident, we also create a \textit{background sentence} that contains information about the birth month, birth year, country of origin (i.e., Denmark or Rest), and sex (SI:~Specification of features and their sources)

\subsubsection{Sentence and Document Structure}
For each resident $r \in \{1,2,3,...,R\}$ in the dataset $\mathcal{D}$, we assemble a chronological sequence of labor and health events. Each life-sequence has a form $S_r = \{s_r^0, s_r^1, s_r^2,..., s_r^{n_r} \}$, where $s_r^i$ is the $i$-th life-event of the $r$-th resident.  

Each event, $s$, contains tokens $v \in \mathcal{V}$ associated with a particular life-event, where $\mathcal{V}$ is a vocabulary of our artificial language. Along with the concept tokens, each event has associated temporal information such as absolute position, age, and segment. $\mathcal{P}$ is a set of possible absolute temporal positions, where $p$ is the number of days passed between the event, $s$, and the \textit{origin point} of 1st January 2008 (the day our dataset starts). If an event happened on the 24th of February 2012, then $p = 1516$. $\mathcal{A}$ is a set of possible age values: $a$ specifies the number of \textit{full} years passed since the person's birthday up until the date of the event, $s$. In terms of the \texttt{life2vec} model, $p$ contextualizes events on a \textit{global} (or universal) time scale, while $a$ contextualizes events on the \textit{individual} timeline.

Lastly, $\mathcal{G}$ is a set of segments. In case two or more events happen on the same day (and thus, share identical age and absolute position), segment information adds additional positional information. We have three distinct segments, and each life-event has an assigned segment value, $g$.  The \texttt{life2vec} model learns the embedding of each segment.

The vocabulary set, $\mathcal{V}$, also includes several special tokens. For example, \texttt{[CLS]} starts a sequence and is later used to encapsulate a dense representation of the sequence. \texttt{[SEP]} token stands between the events, \texttt{[UNK]} substitutes concept tokens that are not in our vocabulary (e.g., tokens that were removed due to the low appearance frequency).

When we refer to the sentence length, $\|s\|$, we refer to the number of the corresponding concept token. The length of every sentence, $s$, varies depending on the type of the event it describes -- health events range from two to three tokens, while labour-events range from three to seven concept tokens. Thus, the final length of the sequence, $\|S_r\|$, is a sum of the length of all the events, plus the number of special tokens such as \texttt{[CLS]} and \texttt{[SEP]}. 

The first sentence in the sequence, $s_r^0$, is a \textit{background sentence} that consists of gender, origin, birth-year, and birth-month tokens. It does not have associated age or absolute time position but does have segment information.

The maximum length of the document is 2560 concept tokens. If the length of the document, $\|S_r\|$, is above the specified limit, we remove earlier events (without removing a background sentence) until we can fit all the tokens of the last sentence (plus, last \texttt{[SEP]}). In case the length of the document is below the limit, we add padding tokens, \texttt{[PAD]}, at the end of the sequence to fill up the empty spaces.

\subsubsection{Data Split}

Finally, we randomly split the dataset (filtered according to (1), (2), (3), and (4) initial requirements) into training, validation, and test sets with a ratio of 70/15/15. The random split is \textit{independent of any features} of the sequence (entirely at random). The global training set has 2~276~460 people, the global validation set has 487~812 people, and the global test set has 487~812 people. We preserve the splits for the finetuning tasks but remove records that do not satisfy specific requirements.

\subsection{Model architecture}
\label{sec:architecture}
The model consists of three components: an embedding layer, a Bert-like encoder \cite{devlin2018bert}, and task-specific decoders. The encoder is a transformer-based model, while decoders are fully-connected neural networks.

\subsubsection{Inputs and Embedding Component}
The first step of the pipeline is to convert life-sequences into dense representations. Given a sequence $S_p$, we look up representations of tokens in the embedding matrix $\mathcal{E}_{\mathcal{V}}: \mathcal{V} \rightarrow \mathbb{R}^d$, where each row of $\mathcal{E}_{\mathcal{V}}$ corresponds to a token in the vocabulary ($d$ is the number of hidden dimensions). Additionally, we look up the segment embedding in the $\mathcal{E}_{\mathcal{G}}: \mathcal{G} \rightarrow \mathbb{R}^d$ matrix. Both $\mathcal{E}_{\mathcal{V}}$ and $\mathcal{E}_{\mathcal{G}}$ matrices are optimized during the model training. To improve the representation of rare concept tokens and the overall isotropy of the concept embedding space \cite{embeddings_shift}, we remove the global mean from each row of the $\mathcal{E}_{\mathcal{V}}$ matrix \cite{embeddings_shift}. That is, each time we look up the token embedding, we subtract the mean.

Regarding age and absolute time positions, we use the \texttt{Time2Vec} \cite{kazemi2019time2vec} method designed to model the linear and periodic progression of time. It introduces two learnable parameters: $\omega$ and $\varphi$. These determine the frequency and phase of periodic functions. The dense representations of age and position are calculated by the following equation, where $z$ specifies the number of dimensions. We initialize two separate sets of time2vec parameters -- one for the age, $\mathcal{T}_{\mathcal{A}} : \mathcal{A} \rightarrow \mathbb{R}^d$, and one for the absolute time position, $\mathcal{T}_{\mathcal{P}}: \mathcal{P} \rightarrow \mathbb{R}^d$. In both cases, we use the cosine function:

\[ \mathcal{T}(x)[z] = \left\{\begin{matrix}
\omega_z x +\varphi_z \phantom{-} & ,\textrm{if } z=0 \phantom{-} \phantom{-} \phantom{-} \\ 
cos(\omega_z x +\varphi_z)& ,\textrm{if } 1 \leq i \leq k.
\end{matrix}\right.
\]

The temporal representation of a sentence, $s_r$, is calculated according to Eq.~\ref{eq:emb_temporal}. Scalars $\alpha$, $\beta$, and $\gamma$ are trainable parameters \cite{bachlechner2020rezero} initialized at a zero value.

\begin{equation}
\label{eq:emb_temporal}
    \mathcal{E}_{temp}(s_r) = \alpha \cdot \mathcal{T}_{\mathcal{A}}(a)  + \beta  \cdot \mathcal{T}_{\mathcal{P}}(p) + \gamma \cdot \mathcal{E}_{\mathcal{G}}(g)
\end{equation}

For each token  $v$ in  $s$, we sum the associated token embedding in $\mathcal{E}_{\mathcal{V}}(v)$ and the temporal embedding of the sentence, $\mathcal{E}_{temp}(s_r^i)$. The input to the \texttt{life2vec} model is a concatenated sequence of these token representations.

\subsubsection{Encoder Component}

Like the original BERT \cite{devlin2018bert}, the \texttt{life2vec}-encoder consists of multiple encoder blocks. Each block processes input representations and passes the results to the next encoder. The architecture of each block is identical and consists of Multi-Head Attention, a Position-wise layer, and two residual connections (SI:~Implementation Details).

The Multi-Head Attention module consists of several attention \textit{heads}, which separately process the input representations. Vanilla BERT \cite{devlin2018bert} uses softmax self-attention heads. Each head takes input representations and transforms these with several dense layers - \textit{query}, \textit{key}, and \textit{value}. These layers output linearly-transformed representations $\mathbf{Q},\mathbf{K}, \mathbf{V}  \in \mathbb{R}^{L \times d}$, where $L$ is the length of the sequence and $d$ is the dimensionality of embeddings.  The contextualised representations are computed as (Note that $\mathbf{}{1}_L$ is a vector of ones with the length of $L$):
\begin{align}
    \displaystyle{Att}(\textbf{Q}, \textbf{K},\textbf{V}) &=  \text{softmax}(\frac{\textbf{Q}\textbf{K}^\text{T}}{\sqrt{d}})\textbf{V} \Leftrightarrow \textbf{D}^{-1}\textbf{A}\textbf{V}, \\ 
    \text{where} \;\textbf{A} &= \text{exp}(\frac{\textbf{Q}\textbf{K}^\text{T}}{\sqrt{d}}), \; \textbf{D}=\text{diag}(\textbf{A}\textbf{1}_L)) 
\end{align}

Softmax attention is unstable for sequences of length more than 512 tokens \cite{beltagy2020longformer}. Therefore, we use softmax attention heads only to model local interactions, i.e., we limit the span of these heads to 38 neighboring tokens.

To capture global interactions, we use Performer-style attention heads \cite{choromanski2020rethinking_performer}, as they can handle longer sequences. Instead of calculating the precise attention matrix $\textbf{A} \in \mathbb{R}^{L \times L}$, Performer-heads approximate it via matrix factorization. Entries of the approximated attention matrix are computed using kernels $\textbf{A'}(i,j) = K(\textbf{q}_i^T, \textbf{k}_j^T)$ (indexes stand for the rows of matrices). The kernel function is defined as $K(\textbf{x},\textbf{y}) = \mathop{{}\mathbb{E}}[\phi(\textbf{x})^T, \phi(\textbf{y})] $, where $\phi(\textbf{u})$ is a random feature map that projects input into the \textit{r}-dimensional space. Random mapping $\phi$ is constrained to contain features that are positive and exactly orthogonal (for details, refer to \cite{choromanski2020rethinking_performer}). If we apply $\phi$ to $\textbf{Q}, \textbf{K}$, we get  $\textbf{Q'}, \textbf{K'} \in \mathbb{R}^{L\times r}$, where $r \ll L$. The attention is now defined as: 
\begin{align}
    \overline{Att}(\textbf{Q}, \textbf{K},\textbf{V}) =   \mathbf{\hat{D}}^{-1}(\textbf{Q'}(\textbf{K'}^T\textbf{V})),  \; \text{where} \; \mathbf{\hat{D}}=\text{diag}(\textbf{Q'}(\textbf{K'}\textbf{1}_L)) 
\end{align}

Each Multi-Head Attention module of the \texttt{life2vec} transformer has four Performer-style attention heads and four Softmax Attention Heads (SI:~Attention Mechanism). The output of these heads is concatenated and transformed with one more dense layer.

The encoder blocks also have a Position-wise Feed-Forward module (\texttt{PFF}). It consists of two fully connected feed-forward layers that apply additional non-linear transformations to each representation: $f_{\mathrm{PFF}}(\mathbf{x}) = \texttt{swish}(\mathbf{x} \mathbf{W}_1 + \mathbf{b}_1)\,\mathbf{W}_2 + \mathbf{b}_2$, where $\texttt{swish}(\mathbf{x}) = \mathbf{x} \cdot \texttt{sigmoid}(\mathbf{x})$ \cite{swish}.

Typically, the output representations of each module add up to the input representations: $\mathbf{y} = \mathbf{x} + f(\mathbf{x})$ \cite{devlin2018bert}, where $f$ is a Multi-Head Attention module or a Position-wise Feed-Forward module. In our work, we use \texttt{ReZero} connections \cite{bachlechner2020rezero}, consisting of a single scalar, $\alpha$. This scalar controls the fraction of information that each layer contributes to the contextualized representations: $\mathbf{y} = \mathbf{x} + \alpha \cdot f(\mathbf{x})$. At the start of training, each $\alpha$ is initialized to zero (meaning that none of the layers contribute. We introduced several modifications to BERT architecture, such as \texttt{ReZero} \cite{bachlechner2020rezero}, \texttt{ScaleNorm} \cite{nguyen2019transformers_scale_norm}, \texttt{Swish} \cite{swish}, and Weight Tying \cite{pappas2018beyond_weight_tying, weight_tying} to speed up the convergence and reduce the size of the model. 

\subsection{Training procedure}
\label{sec:train_procedure}
The training procedure is split into two stages: learning the overall structure of the data (pre-training) and task-specific inference (finetuning).

\subsubsection{Pre-training: Learning Structure of the Data}
During the pre-training stage, \texttt{life2vec} learns embeddings of concept tokens and optimizes the parameters of the encoder component. The training objective consists of two tasks: Masked Language Modeling (MLM)~\cite{vaswani2017attention} and Sequence Order Prediction (SOP).

The Masked Language Modeling task forces the model to learn relations between concept tokens. We randomly choose 30\% of the tokens in the input sequence \cite{wettig2022should}. 80\% of the chosen tokens are substituted with \texttt{[MASK]}, 10\% are unchanged, and 10\% are substituted with random tokens~\cite{devlin2018bert}. We do not mask any special tokens such as \texttt{[CLS]}, \texttt{[SEP]}, \texttt{[PAD]}, or \texttt{[UNK]} (nor do we use them as random tokens). We use altered sequences as inputs to \texttt{life2vec}. Using the contextual output representations of tokens, the model should infer the masked tokens.

The MLM decoder consists of two fully connected layers ($f_1$ and $ f_2$). Each contextual representation, $\mathbf{x}_i$, is transformed via $f_1(\mathbf{x}) = \texttt{tanh}(\mathbf{x}\,\mathbf{W}_1 + \mathbf{b}_1)$, followed by l2-normalisation, $\texttt{norm}(\mathbf{x}) = \mathbf{x} / \| \mathbf{x} \| $. The weights of the final layer, $f_2$, are tied to the embedding matrix, $\mathcal{E}_{\mathcal{V}}$, which is further normalized to preserve only directions \cite{weight_tying}. The resulting scores is scaled by $\alpha$ to sharpen the distribution \cite{nguyen2019transformers_scale_norm}
\begin{equation}
    \mathrm{MLM}(\mathbf{x}) = \alpha \cdot f_2(\,\texttt{norm}(f_1(\mathbf{x})\,)
\end{equation}

For each masked token the model must uncover, the decoder returns the likelihood distribution over the entire vocabulary. The likelihood (in our case) is a product of the scaled cosine distance between the contextualized representation of a token and the original representations of tokens in~$\mathcal{E}_{\mathcal{V}}$~\cite{weight_tying,pappas2018beyond_weight_tying}. 

The Sequence Order Prediction task forces the model to consider the progression of a life-sequence. It is an adapted version of the Next Sentence Prediction task \cite{vaswani2017attention}. Each life-event in the sequence has four attributes: concept tokens, segments, absolute time position, and age. In 10\% of cases, we exchange concept tokens of one life-event with the concept tokens of another life-event (while preserving the positional and temporal information). In half of these cases, the exchange \textit{reverses} the sequence so that 1st life-event exchanges tokens with the last life-event, the second life-2vent exchanges tokens with the second-to-last event, etc. In the other half, we \textit{randomly} pick pairs of life-events to exchange the concept tokens.

The SOP decoder pulls the contextual representation of the \texttt{[CLS]} token from the last encoder layer and passes it through two feed-forward layers to make a final prediction
\begin{equation}
   \mathrm{SOP}(\textbf{x}) = \texttt{ScaleNorm} \left [\, \texttt{swish}(\mathbf{x}\, \textbf{W}_1 + \textbf{b}_1) \,\right ]\,\textbf{W}_2 + \textbf{b}_2
\end{equation}

\subsubsection{Finetuning: Task Specific Training} \label{met:finetuning}

On finetuning, we initialize the model with the optimized parameters from the pre-training stage and assign a new task to the model (i.e., remove the \texttt{MLM} and \texttt{SOP} encoders), which involves initializing a new decoder network.

We evaluate the \texttt{life2vec} model in two settings: Mortality Prediction and Personality Nuances Prediction. For the Mortality Prediction task, we pool the contextualized representation of each token in the sequence (i.e., the output of the last encoder layer) and use a weighted average of these tokens \cite{bahdanau2014neural} to generate Sequence Representations. For the Personality Nuances Prediction Task, we only pool the contextualized representation of the \texttt{[CLS]} token and pass it through a decoder network to make a prediction. The output of the decoder's second-to-last layer is also a Sequence Representation. Refer to SI:~Model Architecture for more details.

The weights of the encoder model are updated during the finetuning. However, deeper encoders have a lower learning rate to avoid `catastrophic forgetting' \cite{sun2019fine}.  We also freeze the parameters of $\mathcal{E}_{\mathcal{V}}$, except for the parameters associated with \texttt{[CLS]}, \texttt{[SEP]} and \texttt{[UNK]} tokens.

\textbf{Mortality Prediction} is a binary classification task. The goal is to infer the mortality likelihood within the next four years after 1st January 2016 (i.e., labels are \textit{alive} and \textit{deceased}).

The crucial aspect of the mortality prediction is the \textit{loss function}. The data we use (see Sec.~\ref{sec:data}) includes people who might have left the country or disappeared before the end of 2020. Hence, we have a handful of \textit{right-censored} outcomes. Using a Cross-Entropy loss would bias the predictions as we do not know the true outcome of all the sequences. Thus, we view the task as a Positive-Unlabeled Learning \cite{wang2021asymmetric} problem. We assume that all negative samples and samples with missing labels make up the unlabeled set, while all positive samples make a Positive-labeled set (see SI:~Implementation Details).

\textbf{Personality Nuances Prediction Task} is an ordinal classification task where labels correspond to the level of agreement with a particular item/statement (five levels). We predict the response to four different items simultaneously. 

Predicting agreement levels poses two technical issues. First, responses are unevenly distributed across possible answers, with a majority choosing non-extreme answers, and second, the level of agreement has an ordinal nature.

We therefore slightly modify the training procedure. To prevent overfitting to the majority class, we employ instance difficulty-based re-sampling \cite{yu2022re}-- samples that are hard to predict would be subsampled with more frequently (SI: Sec.~\ref{app:resampling}). To account for the ordinal and imbalanced nature of the data, we combine three loss functions -- class distance weighted cross-entropy~\cite{polat2022class}, focal loss~\cite{lin2017focal} with label smoothing penalty~\cite{szegedy2016rethinking} (SI:~Loss~Functions), and use a modified softmax function~\cite{kanai2018sigsoftmax}

\subsubsection{Baseline Models}
To evaluate the performance of \texttt{life2vec} on the mortality prediction task, we use six baseline model majority class prediction, random guess, mortality tables, logistic regression, feed-forward neural network, and recurrent neural network (RNN) ~\cite{chung2014empirical_gru, Andreev2002MethodsPF}. We perform a hyperparameter optimization similar to the one we have done for the \texttt{life2vec} model (SI:~Implementation Details).
\begin{itemize}
  \setlength\itemsep{0.1em}
    \item \textbf{Logistic Regression} is a generalized linear regression model. We optimize it using Asymmetrical Cross-Entropy Loss~\cite{wang2021asymmetric} with the ridge penalty and stochastic gradient descent. As an input to the model, we use a counts vector, i.e. how many times each token appears in a sequence over a one-year interval. 
    \item \textbf{Life Tables} is a logistic regression model that uses \textit{only} age and sex as covariates,
    \item \textbf{Feed-forward network} uses the counts vector. It has a similar optimization setting as logistic regression. It has multiple feed-forward layers stacked over each other.
    \item \textbf{RNN model} uses the same input as the \texttt{life2vec} model and the same optimization settings. RNN model outputs the contextual representation of each token, which we then pass through a decoder network (identical to the one in the \texttt{life2vec}'s one).
\end{itemize}

These models work with the same data (i.e., batches of data are identical) and the same optimization settings.

For the Personality Nuances Prediction Task, we use a random guess and the RNN model. The \texttt{life2vec} model pools only the \texttt{[CLS]} representation from the decoder; however, with the RNN model, we pool all the contextual representations from the RNN (this way, we improve the performance of the RNN-based model). 

\subsubsection{Data Augmentation}
To stabilize the performance of the \texttt{life2vec} model, we introduce several data augmentation strategies.  It was an essential part of the training procedure and helped boost the performance of \texttt{life2vec} and baseline models. The augmentation techniques include subsampling sentences and tokens, adding noise to the temporal information, and masking the background sentence (SI:~Implementation Details).

\subsection{Interpretability}
\label{subchap:methods_interpretability}
To provide the local interpretability, we use the Gradient-based Saliency score with L2-normalisation \cite{atanasova2020diagnostic_saliency, bastings2020elephant, ding2019saliency_smooth}.  The saliency score highlights the sensitivity of the output with respect to each input token, i.e., the higher the sensitivity score, the more the output changes if we change the token representation (SI:~Implementation Details).

\textbf{TCAV}. Gradient-based Saliency is unreliable when we want to see the global sensitivity of a model towards certain concepts (on a global scale). The person-summaries (provided by the \texttt{life2vec}) form a complex multidimensional space. Dimensions of this space do not necessarily have human-interpretable meaning.  Thus, we use Testing with Concept Activation Vectors (TCAV)~\cite{kim2018interpretability} to estimate the overall sensitivity. 

We define a high-level concept as a subsample of life-sequences that share specific attributes (such as ``individual has an F-diagnosis in the sequence''). We can take sequence representations of this subsample and train a linear classifier to discriminate between sequences in concept and random subsamples. The normal to the decision hyperplane is a concept direction. To calculate the TCAV scores, we rely on the following procedure \cite{kim2018interpretability} (SI:~Implementation Details). 

\subsection{Evaluation of the Concept Space}
\label{subchap:test_concept_space}

While the structure of the Concept Space (Fig.~\ref{fig:embeddings}) seem reasonable under manual inspection, we provide further statistical proof for the robustness of the embedding. 

To demonstrate the \textbf{robustness of the concept space}, we used randomization tests \cite{kriegeskorte2008representational}. Here we test if the model preserves the distances between pairs of concept tokens given different dataset splits.

We trained three models with identical architecture. Each model had a different random initialization and was trained on a unique subset of the training data for ten epochs.

Further, we extracted the trained concept embeddings and calculated the cosine distances between each concept for each model separately (we refer to these matrices as $\mathcal{M}_1$, $\mathcal{M}_2$, and $\mathcal{M}_3$). We also obtained the distance matrices based on the randomly initialized embeddings and on the permuted version of $\mathcal{M}_1$ (referred to as $\mathcal{R}$ and $\mathcal{P}$, respectively).

To prove that our embedding spaces preserve structure/distances, we test whether two matrices are correlated. To perform the comparison, we use Randomisation Test described in \cite{kriegeskorte2008representational}.  For each pair of matrices, we permute columns and rows of the first matrix and calculate the correlation between permuted and the second matrix. We run the procedure 1~000 times. As a result, we get a distribution of correlation coefficients under the null hypothesis that there is no relationship between the two matrices. Suppose the correlation between the initial matrices is higher than the randomized one (falls above $95$-th quantile of a distribution); in that case, we can indeed assume that the two are similar and, thus, distances between concepts are similar. To account for the multiple testings $(\mathcal{M}_1,\; \mathcal{M}_2)$, $(\mathcal{M}_1, \mathcal{M}_2)$, $(\mathcal{M}_2, \mathcal{M}_3)$, $(\mathcal{M}_1, \mathcal{R})$,  $(\mathcal{M}_1, \mathcal{P})$ we use Benjamini–Hochberg procedure \cite{thissen2002quick}.  We reject the null hypothesis in the first three pairs with $p \approx 3\mathrm{e}{-4}$ in all cases and accept the null hypothesis in case of the random-comparison case ($p\approx .76$) and permuted-comparison case ($p\approx .37$). 

Our evaluation shows that the concept space converges to a similar space structure for each subset of a dataset.

\textbf{Hubness of the Concept Space}. The embedding spaces produced by ML models often degenerate due to the presence of the low-frequency tokens \cite{liang2021learning, embeddings_shift}. The model places tokens along a similar direction, leading to less meaningful representations. The presence of hubs (tokens with an abnormal number of neighbors) is a proposed proxy for the degeneration of the embedding space \cite{mu2017all} (aka anisotropy).

To identify hubs in the embedding matrix, $\mathcal{E}_{\mathcal{V}}$, we found the five closest neighbors of each node based on cosine similarity and used the resulting adjacency matrix to create a directed graph. Hubs can be identified by counting the incoming edges, which are the tokens with a large number of incoming edges. However, we did not find any hubs (i.e., nodes with an abnormally large number of incoming connections). The \texttt{[PAD]} token has the highest number of incoming connections (i.e., 49 links), \texttt{[CLS]} (40 links),  \texttt{[SEP]} (39 links), followed by \texttt{[Female]} (25), \texttt{[Male]} (24) -- the token with the most incoming edges is neighbor to less than 2\% of tokens. Thus, we do not find proof of a degenerated concept space.

In summary, \texttt{life2vec} produces a meaningful and robust representation of the building blocks of our synthetic language.

\subsubsection{Evaluation Metric for Task-Specific Settings}
\label{sec:methods_evaluation}
Since \textbf{Mortality Prediction Task} is a PU-Learning task, we cannot use standard metrics to evaluate the model without introducing a bias \cite{ramola2018estimating}. We evaluate models using the \textit{Corrected} Matthews Correlation Coefficient, C-MCC (see \cite{ramola2018estimating} for details), as well as the Area-Under the Lift (AUL) \cite{jiang2020improving}. We also provide the corrected balanced accuracy score and corrected F1-score (SI:~Evaluation Details).

We use AUL for the model optimization as suggested in \cite{jiang2020improving}. i.e., early stopping. AUL can be interpreted as the ``\textit{probability of correctly ranking a random positive sample versus a random negative sample}'' \cite{huang2020aul}.

We use bootstrapping to estimate the confidence intervals for the corrected C-MCC score. 

For the \textbf{Personality Nuances Prediction Task}, we use Cohens's Quadratic Kappa (CQK) score to terminate the training (when the score decreases on the validation set) \cite{polat2022class}. We also use CQK to evaluate and compare models. 

\end{small}
\newpage

\bibliographystyle{unsrt}
\bibliography{sn-bibliography}

\newpage
\appendix
\setcounter{table}{0}
\setcounter{figure}{0}                      
\setcounter{equation}{0}                      

\renewcommand{\thetable}{A\arabic{table}}
\renewcommand{\thefigure}{A.\arabic{figure}} 
\renewcommand{\theequation}{A.\arabic{equation}} 

\newpage
\section{Definitions}

\begin{table}[htbp]\caption{Table of Notations}
\centering 
\begin{tabular}{r c p{10cm} }
\toprule
$\texttt{PFF}$ & $\triangleq$ & Position-wise feed forward module\\

$\texttt{MLM}$ & $\triangleq$ & Masked Language Model\\
$\texttt{SOP}$ & $\triangleq$ & Sequence Order Prediction\\

$S_r$ & $\triangleq$ & a life-sequence of a $r$-th resident \\
$s_r^i$ & $\triangleq$ & $i$-th event in a life-sequence of a resident $r$\\
$L$ & $\triangleq$ & maximum length of a sequence\\
$d$ & $\triangleq$ & number of hidden dimensions\\

$\mathcal{E}_{\mathcal{V}}$ & $\triangleq$ & embedding matrix of concepts\\ 

$\mathcal{E}_{\mathcal{G}}$ & $\triangleq$ & embedding matrix of segments\\ 
$\mathcal{T}_{\mathcal{A}}$ & $\triangleq$ & time2vec embedding of the age\\ 
$\mathcal{T}_{\mathcal{P}}$ & $\triangleq$ & time2vec embedding of the absolute position\\ 

$\textbf{ff}_i$ & $\triangleq$ & $i$-th fully connected layer\\

$\textbf{W}_i$ & $\triangleq$ & weight matrix of the $i$-th layer\\
$\textbf{b}_i$ & $\triangleq$ & bias vector of the $i$-th layer\\
$g$ & $\triangleq$ & trainable  parameter\\
$\textbf{A}$ & $\triangleq$ & Attention score matrix\\

$\textbf{1}_L$ & $\triangleq$ & vector of \textbf{1}s with the length of $L$\\

\multicolumn{3}{c}{}\\
\multicolumn{3}{c}{\underline{Functions}}\\
$\texttt{Norm}(x)$ & $=$ & $\frac{x}{\| x \|}$\\

$\texttt{ScaleNorm}(x)$ & $=$ & $g \cdot \frac{x}{\| x \|}$ \quad \cite{nguyen2019transformers_scale_norm}\\
$\texttt{sigmoid}(x)$ & $=$ & $\frac{1}{1 + e^{-1}}$\\

$\texttt{swish}(x)$ & $=$ & $x \cdot \texttt{sigmoid}(x)$   \quad \cite{swish}\\

$\texttt{SigSoftmax}(x_i)$ & $=$ & $\frac{\texttt{exp}(x_i) \cdot \texttt{sigmoid}(x_i)}{\sum \texttt{exp}(x_j) \cdot \texttt{sigmoid}(x_j)}$   \quad \cite{kanai2018sigsoftmax}\\

\bottomrule
\end{tabular}
\label{tab:TableOfNotationForMyResearch}
\end{table}

\newpage
\section{Evaluation Details}

\begin{table}[h]
\centering
\caption{Corrected Matthew's Correlation Coefficient (MCC) and Area under the Lift (AUL) on the Mortality Prediction Task.  95\%-Confidence intervals for the MCC based on the stratified bootstrapping. In both cases, the higher value is preferred. Model Size specifies the number of trainable parameters, we performed a hyperparameter tuning on RNN, FFNN, and Logistic Regression.}
\label{tab:eos_metric_a}
\begin{tabular}{@{}lllllc@{}}
\toprule
\multicolumn{1}{c}{\textbf{Model}} &
  \multicolumn{1}{c}{\textbf{MCC, 95\%-CI}} &
  \multicolumn{1}{c}{\textbf{AUL}} &
  \multicolumn{1}{c}{\textbf{Accuracy, 95\%-CI}} &
  \multicolumn{1}{c}{\textbf{F1-Score, 95\%-CI}} &
  \multicolumn{1}{c}{\textbf{Model Size}} \\ \midrule
\textbf{L2V}   & \textbf{0.413 {[}0.410, 0.422{]}} & \textbf{0.845} & 0.788 {[}0.782, 0.794{]} & 0.443 {[}0.435, 0.451{]} & 8.4m\\
RNN-GRU        & 0.369 {[}0.361, 0.378{]}          & 0.834          & 0.778 {[}0.771, 0.783{]} & 0.395 {[}0.389, 0.402{]} & 1.5m\\
FFNN           & 0.340 {[}0.332, 0.348{]}          & 0.822          & 0.768 {[}0.762, 0.774{]} & 0.345 {[}0.339, 0.350{]} & 8.4m\\
Logistic Reg   & 0.149 {[}0.142, 0.155{]}          & 0.735          & 0.639 {[}0.633, 0.645{]} & 0.201 {[}0.198, 0.204{]} & 2.0k\\
Life Tables    & 0.059 {[}0.051, 0.066{]}          & 0.650          & 0.555 {[}0.548, 0.562{]} & 0.161 {[}0.158, 0.164{]} &  3\\
Random         & -0.005 {[}-0.011, 0.002{]}        & 0.497          & 0.496 {[}0.489, 0.503{]} & 0.132 {[}0.128, 0.135{]} & - \\
Majority Class & 0.0                               & 0.497          & 0.5                      & -                        & - \\ \bottomrule
\end{tabular}
\end{table}

\begin{table}[h]
\centering
\caption{Corrected Matthew's Correlation Coefficient (MCC) and Area under the Lift (AUL) on the \textit{Emigration} Prediction Task.  95\%-Confidence intervals for the MCC based on the stratified bootstrapping. In both cases, the higher value is preferred.}
\label{tab:eos_metric_b}
\begin{tabular}{@{}lllllc@{}}
\toprule
\multicolumn{1}{c}{\textbf{Model}} &
  \multicolumn{1}{c}{\textbf{MCC, 95\%-CI}} &
  \multicolumn{1}{c}{\textbf{AUL}} &
  \multicolumn{1}{c}{\textbf{Accuracy, 95\%-CI}} &
  \multicolumn{1}{c}{\textbf{F1-Score, 95\%-CI}} &
  \multicolumn{1}{c}{\textbf{Model Size}} \\ \midrule
\textbf{L2V}   & \textbf{0.168 {[}0.159, 0.177{]}} & \textbf{0.802} & 0.731 {[}0.719, 0.744{]} & 0.130 {[}0.125, 0.134{]} & 8.4m\\
RNN-GRU        & 0.144 {[}0.136, 0.151{]}          & 0.786          & 0.714 {[}0.702, 0.726{]} & 0.106 {[}0.103, 0.110{]} & 1.5m\\
Random         & 0.000 {[}-0.001, 0.009{]}         & 0.504          & 0.499 {[}0.486, 0.513{]} & 0.052 {[}0.049, 0.055{]} & - \\
Majority Class & 0.0                               & 0.504          & 0.5                      & -                        & - \\ \bottomrule
\end{tabular}
\end{table}

\newpage
\section{Specification of features and their sources}
\label{app:variables}

\begin{table}[h]
\centering
\caption{Specification of features and their sources. }
\label{tab:variables}
\scalebox{0.6}{
\begin{tabular}{rllll}
\multicolumn{1}{c}{\textbf{Type}} & \multicolumn{1}{c}{\textbf{Feature}} & \multicolumn{1}{c}{\textbf{Source}} & \multicolumn{1}{c}{\textbf{\# Categories}} & \multicolumn{1}{c}{\textbf{Encoding}} \\
\multirow{4}{*}{\textbf{\begin{tabular}[c]{@{}r@{}}Background \\ Information\end{tabular}}} & Sex & KOEN & 2 binary & Male, Female \\
 & Birth Month & FOED\_DAG & 12 & Jan-Feb \\
 & Birth Year & FOED\_DAG & 45 & 1946-1991 \\
 & Country of Origin & OPR\_LAND & 2 binary & National or International \\
\multirow{9}{*}{\textbf{\begin{tabular}[c]{@{}r@{}}Labour \\ Records\end{tabular}}} & Municipality of Residence & BOPAEL\_KOM\_KODE & 97 & Danish municipality codes \\
 & Tax Bracket & ATP\_BIDRAG\_SATS\_KODE & 6 & based on DST definitions \\
 & Income Level & BREDT\_LOEN\_BELOEB & 100 & Quantile-based \\
 & Labour Force Status & SOC\_STATUS\_KODE & 35 & based on DST definitions \\
 & Labour Force Status (Modification) & TILSTAND\_KODE\_AMR & 58 & based on DST definitions \\
 & Labour-Force-Interval & TILSTAND\_LAENGDE\_ARR & 10 & Quantile based \\
 & Industry Area (Company) & ARB\_HOVED\_BRA\_DB07 & 290 & Danish Industry Classification System \\
 & Job type & DISCO\_KODE & 359 & International Standard Classification of Occupations \\
 & Enterprise Type (Company) & ARB\_SEKTORKODE & 15 & European System of Accounts \\
\multirow{3}{*}{\textbf{\begin{tabular}[c]{@{}r@{}}Health \\ Records\end{tabular}}} & Diagnosis & C\_ADIAG & 704 & ICD-10 \\
 & Urgency & C\_INDM & 3 & Urgent, Non-Urgent, Emergency \\
 & Patient Type & C\_PATTYPE & 2 & In-, out- patient
\end{tabular}}
\end{table}

\newpage
\section{Hyperparameter Optimization}
\begin{table}[ht!]
\centering
\caption{Hyperparameter optimisation for the life2vec model. We use Bayesian search to find the optimal configuration of the parameters. The overall perplexity is calculated as a weighted sum of perplexities generated by the MLM task (0.7) and Sequence Order Prediction task (0.3). We train each model for 5 epochs and then pick 6 models with the lowest scores. Lastly, we train these six models for 30 epochs and choose the one with the lowest perplexity score on the validation set. Model Nr. 3 is the final configuration of the life2vec model.  }
\label{tab:transformer_optimisation}
\scalebox{0.6}{
\begin{tabular}{lrrrrrrrr}
\multicolumn{1}{c}{\textbf{ID}} & \multicolumn{1}{c}{\textbf{\begin{tabular}[c]{@{}c@{}}Overall \\ Perplexity\end{tabular}}} & \multicolumn{1}{c}{\textbf{\begin{tabular}[c]{@{}c@{}}Hidden \\ Size\end{tabular}}} & \multicolumn{1}{c}{\textbf{\# encoders}} & \multicolumn{1}{c}{\textbf{\# heads}} & \multicolumn{1}{c}{\textbf{\begin{tabular}[c]{@{}c@{}}\# local \\ heads\end{tabular}}} & \multicolumn{1}{c}{\textbf{\begin{tabular}[c]{@{}c@{}}FF Hidden \\ Size\end{tabular}}} & \multicolumn{1}{c}{\textbf{\begin{tabular}[c]{@{}c@{}}\# random\\  features\end{tabular}}} & \multicolumn{1}{c}{\textbf{\begin{tabular}[c]{@{}c@{}}Local \\ Window \\ Size\end{tabular}}} \\
0 & 1.870 & 332 & 13 & 4 & 3 & 996 & 242 & 52 \\
1 & 1.843 & 238 & 13 & 14 & 4 & 1586 & 263 & 32 \\
2 & 1.859 & 208 & 13 & 8 & 3 & 2235 & 413 & 93 \\
\textit{\textbf{3}} & \textit{\textbf{1.835}} & \textit{\textbf{280}} & \textit{\textbf{5}} & \textit{\textbf{10}} & \textit{\textbf{7}} & \textit{\textbf{2210}} & \textit{\textbf{436}} & \textit{\textbf{38}} \\
4 & 1.925 & 80 & 11 & 10 & 6 & 1355 & 153 & 40 \\
5 & 1.908 & 96 & 12 & 8 & 1 & 1790 & 360 & 41 \\
6 & 1.881 & 184 & 6 & 4 & 1 & 709 & 469 & 19 \\
7 & 1.857 & 196 & 5 & 14 & 11 & 1124 & 326 & 74 \\
8 & 1.838 & 228 & 14 & 6 & 1 & 1605 & 490 & 65 \\
9 & 1.859 & 208 & 12 & 4 & 1 & 2135 & 77 & 114 \\
10 & 1.846 & 210 & 8 & 14 & 5 & 1615 & 356 & 49 \\
11 & 1.889 & 312 & 8 & 13 & 12 & 1991 & 138 & 165 \\
12 & 1.867 & 216 & 10 & 4 & 2 & 1839 & 223 & 99 \\
13 & 1.886 & 154 & 12 & 14 & 1 & 1532 & 97 & 102 \\
14 & 1.916 & 70 & 8 & 14 & 5 & 2074 & 410 & 41 \\
\textbf{15} & \textbf{1.829} & \textbf{270} & \textbf{6} & \textbf{10} & \textbf{1} & \textbf{1964} & \textbf{275} & \textbf{99} \\
16 & 1.848 & 242 & 4 & 11 & 10 & 2386 & 120 & 80 \\
17 & 1.884 & 168 & 7 & 12 & 7 & 1702 & 66 & 229 \\
18 & 1.865 & 336 & 4 & 12 & 10 & 2432 & 512 & 14 \\
19 & 1.848 & 336 & 4 & 8 & 4 & 2560 & 512 & 4 \\
20 & 1.889 & 162 & 6 & 9 & 4 & 1214 & 188 & 137 \\
21 & 1.886 & 220 & 9 & 11 & 8 & 2482 & 271 & 188 \\
22 & 1.846 & 336 & 7 & 12 & 4 & 2560 & 512 & 68 \\
23 & 1.873 & 336 & 10 & 8 & 7 & 2560 & 512 & 4 \\
24 & 1.878 & 310 & 5 & 5 & 2 & 2049 & 241 & 108 \\
25 & 1.867 & 322 & 6 & 14 & 9 & 882 & 282 & 150 \\
26 & 2.036 & 270 & 9 & 10 & 0 & 2322 & 192 & 208 \\
27 & 1.857 & 288 & 7 & 6 & 1 & 1964 & 325 & 154 \\
28 & 2.014 & 144 & 9 & 6 & 0 & 1906 & 395 & 198 \\
29 & 1.867 & 242 & 10 & 11 & 1 & 2285 & 158 & 140 \\
30 & 1.925 & 301 & 8 & 7 & 5 & 1380 & 296 & 183 \\
31 & 1.944 & 120 & 5 & 12 & 5 & 2391 & 361 & 130 \\
32 & 1.851 & 252 & 7 & 9 & 1 & 1771 & 209 & 254 \\
33 & 1.878 & 294 & 10 & 7 & 5 & 2506 & 316 & 167 \\
34 & 1.873 & 300 & 5 & 10 & 9 & 2124 & 429 & 82 \\
35 & 1.857 & 275 & 6 & 11 & 10 & 2019 & 452 & 118 \\
36 & 1.946 & 261 & 4 & 9 & 5 & 1304 & 174 & 230 \\
37 & 1.878 & 312 & 5 & 13 & 7 & 1713 & 376 & 169 \\
\textbf{38} & \textbf{1.827} & \textbf{253} & \textbf{6} & \textbf{11} & \textbf{9} & \textbf{2439} & \textbf{133} & \textbf{120} \\
39 & 1.916 & 171 & 5 & 9 & 3 & 2168 & 297 & 254 \\
40 & 1.857 & 273 & 7 & 7 & 3 & 1904 & 250 & 89 \\
\textbf{41} & \textbf{1.835} & \textbf{286} & \textbf{7} & \textbf{13} & \textbf{9} & \textbf{1446} & \textbf{207} & \textbf{138} \\
42 & 1.859 & 234 & 4 & 9 & 4 & 2352 & 315 & 25 \\
43 & 1.838 & 286 & 8 & 11 & 6 & 1213 & 346 & 55 \\
44 & 1.848 & 260 & 6 & 13 & 12 & 556 & 114 & 116 \\
45 & 1.881 & 238 & 9 & 14 & 10 & 2560 & 106 & 133 \\
46 & 2.039 & 294 & 4 & 14 & 0 & 2560 & 236 & 53 \\
47 & 2.031 & 336 & 6 & 14 & 0 & 1912 & 180 & 256 \\
48 & 1.870 & 108 & 13 & 12 & 10 & 1388 & 98 & 103 \\
49 & 1.952 & 70 & 7 & 10 & 10 & 1454 & 334 & 4 \\
50 & 1.938 & 70 & 4 & 14 & 12 & 1096 & 497 & 4 \\
51 & 1.933 & 66 & 4 & 3 & 2 & 877 & 231 & 256 \\
52 & 1.838 & 294 & 13 & 14 & 8 & 2412 & 91 & 4 \\
\textbf{53} & \textbf{1.829} & \textbf{336} & \textbf{14} & \textbf{14} & \textbf{2} & \textbf{1785} & \textbf{105} & \textbf{4} \\
54 & 1.862 & 252 & 10 & 14 & 2 & 2312 & 116 & 4 \\
\textbf{55} & \textbf{1.832} & \textbf{336} & \textbf{14} & \textbf{14} & \textbf{2} & \textbf{2096} & \textbf{64} & \textbf{256} \\
\end{tabular}}
\end{table}

\begin{table}[]
\centering
\caption{Hyperparameter optimization for the \texttt{RNN-GRU} (Mortality Prediction). We use Bayesian search to find the optimal configuration of the parameters. We optimize parameters that were the most sensitive with respect to the performance of the model (determined manually). We pick the model with the highest AUL score (on the validation set after 5 epochs). Model 18 has the most optimal configuration of the hyperparameters.}
\label{tab:optimization_rnn}
\begin{tabular}{lccccl}
\multicolumn{1}{c}{\textbf{ID}} & \textbf{Hidden Size} & \textbf{\# layers} & \textbf{Dropout, \%} & \textbf{Bidirectional} & \multicolumn{1}{c}{\textbf{AUL}} \\
1 & 370 & 1 & 0.16 & False & 0.7898 \\
2 & 292 & 7 & 0.49 & False & 0.7824 \\
3 & 155 & 7 & 0.02 & False & 0.7902 \\
4 & 137 & 5 & 0.39 & False & 0.7898 \\
5 & 525 & 5 & 0.01 & False & 0.7903 \\
6 & 361 & 3 & 0.48 & False & 0.7865 \\
7 & 646 & 3 & 0.35 & False & 0.7865 \\
8 & 372 & 2 & 0.26 & True & 0.7902 \\
9 & 260 & 7 & 0.09 & False & 0.7897 \\
10 & 408 & 1 & 0.48 & True & 0.7890 \\
11 & 367 & 2 & 0.37 & False & 0.7890 \\
12 & 415 & 1 & 0.37 & True & 0.7897 \\
13 & 267 & 1 & 0.40 & True & 0.7910 \\
14 & 64 & 4 & 0.28 & True & 0.7910 \\
15 & 64 & 1 & 0.50 & False & 0.7873 \\
16 & 304 & 4 & 0.00 & True & 0.7893 \\
17 & 768 & 8 & 0.00 & True & 0.7888 \\
\textbf{18} & \textbf{256} & \textbf{3} & \textbf{0.27} & \textbf{True} & \textbf{0.7912} \\
19 & 710 & 2 & 0.04 & False & 0.7888 \\
20 & 490 & 8 & 0.26 & False & 0.7375 \\
21 & 268 & 8 & 0.21 & False & 0.7902 \\
22 & 657 & 8 & 0.18 & True & 0.7728 \\
23 & 248 & 8 & 0.14 & False & 0.7882 \\
24 & 520 & 1 & 0.10 & True & 0.7893
\end{tabular}
\end{table}

\begin{table}[]
\centering
\caption{Hyperparameter optimization for the Feedforward neural network (Mortality Prediction). We use Bayesian search to find the optimal configuration of the parameters. We optimize parameters that were the most sensitive with respect to the performance of the model (determined manually). We pick the model with the highest $AUL$ score (on the validation set after 5 epochs). Model 7 has the most optimal configuration of the hyperparameters.}
\label{tab:optimization_tab}
\begin{tabular}{llllll}
\textbf{ID} & \textbf{Hidden Size} & \textbf{Layers} & \textbf{Dropout} & \textbf{LR} & \textbf{AUL} \\
1 & 370 & 1 & 0.16 & 0.00039 & 0.7260 \\
2 & 292 & 7 & 0.49 & 0.00043 & 0.6950 \\
3 & 155 & 7 & 0.02 & 0.00378 & 0.7230 \\
4 & 137 & 5 & 0.39 & 0.00374 & 0.7199 \\
5 & 525 & 5 & 0.01 & 0.00303 & 0.7234 \\
6 & 361 & 3 & 0.48 & 0.00431 & 0.7196 \\
\textbf{7} & \textbf{646} & \textbf{3} & \textbf{0.35} & \textbf{0.00017} & \textbf{0.7290} \\
8 & 372 & 2 & 0.26 & 0.00503 & 0.7216 \\
9 & 260 & 7 & 0.09 & 0.00296 & 0.7203 \\
10 & 408 & 1 & 0.48 & 0.00595 & 0.7223 \\
11 & 64 & 5 & 0.19 & 0.01000 & 0.7193 \\
12 & 64 & 1 & 0.39 & 0.01000 & 0.7239 \\
13 & 768 & 3 & 0.25 & 0.00001 & 0.6914 \\
14 & 768 & 2 & 0.00 & 0.00171 & 0.7250 \\
15 & 768 & 3 & 0.00 & 0.00109 & 0.7258 \\
16 & 599 & 7 & 0.00 & 0.00718 & 0.7196 \\
17 & 665 & 1 & 0.47 & 0.00874 & 0.7171 \\
18 & 768 & 2 & 0.24 & 0.00041 & 0.7246 \\
19 & 363 & 3 & 0.35 & 0.00736 & 0.7185 \\
20 & 462 & 2 & 0.24 & 0.00065 & 0.7255 \\
21 & 110 & 2 & 0.21 & 0.00700 & 0.7232 \\
22 & 592 & 3 & 0.38 & 0.00866 & 0.7190 \\
23 & 511 & 2 & 0.20 & 0.00273 & 0.7236 \\
24 & 155 & 2 & 0.21 & 0.00244 & 0.7250
\end{tabular}
\end{table}
\newpage
\section{Implementation Details}
\subsection{Ecnoder-Decoder Architecture} \label{app:architecture}
\begin{figure}[ht!]
    \centering
    \label{fig:architecture_viz}
    \makebox[\textwidth][c]{\includegraphics[width=0.8\textwidth]{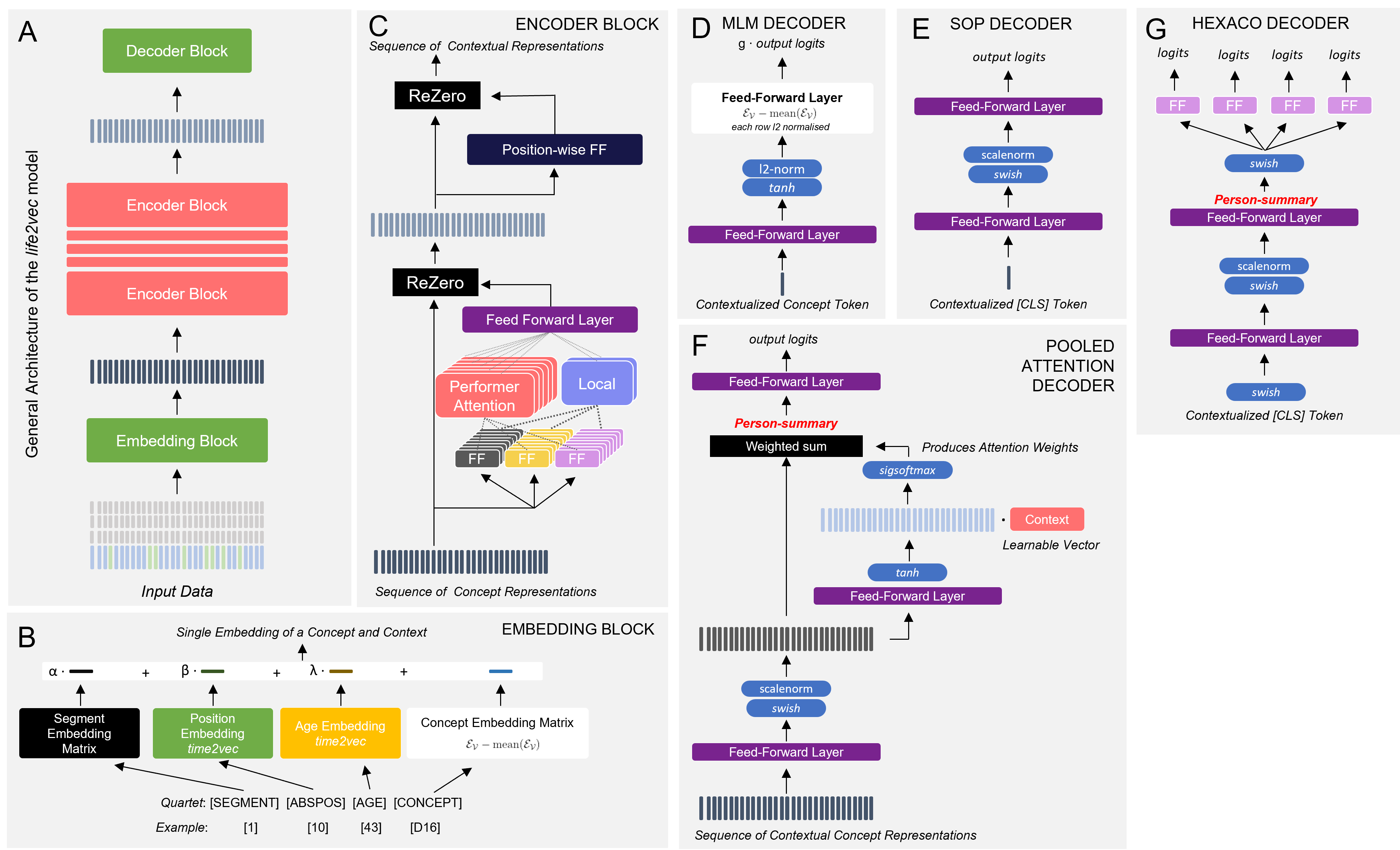}}
    \caption{\textbf{Architecture of \texttt{life2vec}}. (\textbf{A}) The overall structure of the model. The life-sequence is passed to the embedding block and every encoder block. The implementation of a decoder block depends on the task. Each block before the decoder outputs a contextualized representation of concept tokens. (\textbf{B}) Each concept token in a sequence (together with the positional information) passes through the embedding block. It merges positional and concept information. Age and Absolute Position embeddings are calculated via \texttt{time2vec} \cite{kazemi2019time2vec}, while segment and concept embeddings are stored in a lookup matrix. To retrieve the concept embedding, we lookup the corresponding embedding in a matrix $\mathcal{E}_{\mathcal{V}}$ and then remove the mean of $\mathcal{E}_{\mathcal{V}}$ (without changing the representation in an actual $\mathcal{E}_{\mathcal{V}}$. Positional representations are weighted and added to the representation of a concept. (\textbf{C}) The encoder block takes the life-sequence representation from the previous block. The sequence is passed through the attention layer (Multi-head Performer and Local Attention). The result of the attention layer is added to life-sequence representation via \texttt{ReZero} gate. Further, the life-sequence representation is passed through the Position-Wise Feed Forward Layer. The result is added back to the life-sequence representation via another \texttt{ReZero} gate. (\textbf{D}) Masked Language Model Decoder pools (separately) contextual representation of masked concept tokens. MLM Decoder and the Sequence Order Prediction Decoder (\textbf{E}) are used during the pre-training. The last feed-forward layer copies weights from the matrix $\mathcal{E}_{\mathcal{V}}$. We remove the mean of  $\mathcal{E}_{\mathcal{V}}$ from each row and then apply the l2-norm. Thus, the output logits are calculated as a dot product between the contextual representation of a concept and each row of standardized $\mathcal{E}_{\mathcal{V}}$ matrix \cite{nguyen2019transformers_scale_norm}. (\textbf{E}) The Sequence Order Prediction decoder pools the contextual representation of \texttt{[CLS]} concept (which is always placed first in the sequence). (\textbf{F}) The pooled attention decoder is used during the finetuning. It pools the life-sequence from the last encoder block. After passing through the first feed-forward layer, it uses life-sequence representations to compute attention weights (right-hand side). It uses context (a trainable vector) to calculate the importance of a concept at a position $i$. Attention weights are then used to calculate the weighted average of concepts in the life-sequence, or \textit{person-summary}. (\textbf{G}) HEXACO Decoder pools the contextual representation of \texttt{[CLS]} and computes the logits per each personality item.}
\end{figure}

\subsection{Data Augmentation}
\label{app:augmentation}

We use data augmentation during the pre-training and fine-tuning stages. These techniques ensure robust performance and better generalizability of the model.

\textbf{Sequence Downsampling}. We randomly pick a life-sequence and randomly remove up to 50\% of life-events.

\textbf{Temporal Noise} As labor-events mainly occur on the last day of a month, we smooth the distribution of the absolute time. We randomly pick a life-sequence and alter the absolute time by injecting noise, $\mathcal{U}(-5,5)$, into each life-event. 

\textbf{Background Masking}. We randomly pick a life-sequence and mask the background information (i.e., sex, origin, and birthday) with the \texttt{[UNK]} token.

\textbf{Token Downsampling}. We randomly pick a life-sequence and randomly remove tokens from the life-events. This procedure does not affect \texttt{[CLS]} and \texttt{[SEP]} tokens.

The augmentation procedures are independent of each other. Thus, some sequences might be altered by multiple procedures—the order of application: Sequence Downsampling, Temporal Noise, Background Masking, and Token Downsampling. 

\subsection{Interpretability}
\label{app:interpretability}
\textbf{Attention Score} When we use the Pooled Attention Decoder (see SI~Sec.~\ref{app:architecture}, we can extract the attention weights associated with each concept token in a sequence. 

\textbf{Saliency score} indicates the degree of change (and is directly connected to the partial derivatives). The contribution is calculated by back-propagating through the network, starting from the output score toward each token in the sequence. The higher the gradient associated with a token, the higher the contribution towards the predicted values since small changes in a particular concept token embedding would lead to a higher degree of change in the output. To achieve robust scores and minimize the noise associated with the gradient descent, we use \texttt{SmoothGrad} implementation \cite{ding2019saliency_smooth} of saliency. $x_i$ is an embedding of a token, $f(x_{1:n})$ is the output of the model, $n$ is the number of noisy samples. 

\[S(x_i) = \frac{1}{n}\sum_{n}  \left \| \; \frac{\partial }{\partial x_i} f(x_{1:n})  x_i + e \; \right \|_2, \quad e \sim \mathcal{N}(0, \sigma^{2}) \]

\begin{figure}[h]
    \centering
    \makebox[\textwidth][c]{\includegraphics[width=0.8\textwidth]{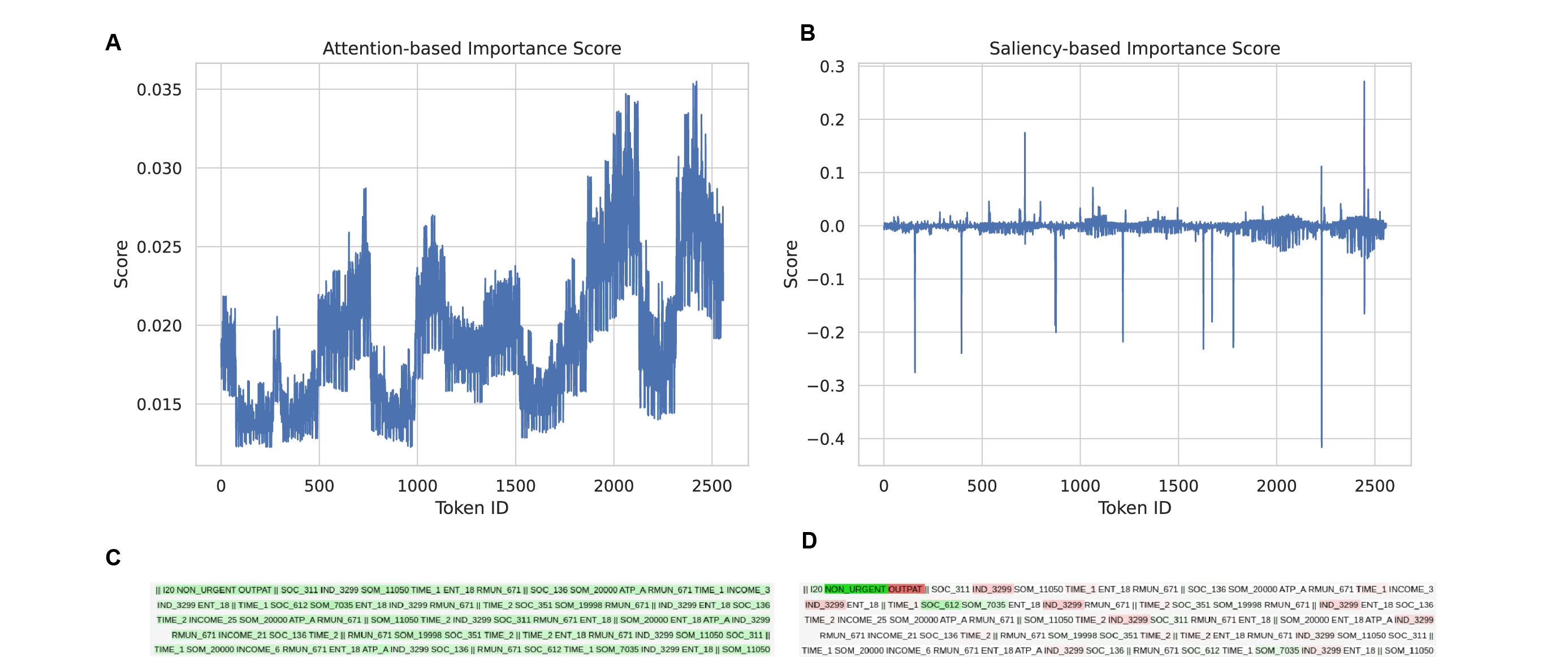}}
    \caption{Visualisation of Attention and Saliency Scores for a specific sequence (only small piece of a sequence is shown). This sequence belongs to a person who survived the four-year interval, the model assigned the probability of an early mortality of $0.23$. (\textbf{A} - \textbf{B}) Attention and Saliency score per each token in the sequence. (\textbf{C}) Part of a life-sequence, each word is colored based on the assigned Attention score. (\textbf{D})  Part of a life-sequence, each word is colored based on the assigned Saliency score. Saliency scores provide more comprehensive importance score: red stands for negative contribution, and green stands for the positive contribution. $\texttt{OUTPAT}$ (outpatient visit) lowers the probability of early mortality, as well as $\texttt{IND\_3299}$ (Manufacturing activities). However, $\texttt{SOC\_612}$ (sick leave) increases the probability of death. The comparison between the Attention and Saliency scores supports the claim that Saliency is more meaningful importance score \cite{atanasova2020diagnostic_saliency}.}
    \label{fig:attributionviz}
\end{figure}

Attention scores and Saliency Scores help us understand how big of a contribution each token towards the final prediction. The example is shown in a Fig.~\ref{fig:attributionviz}.

\textbf{Discovery of the concept directions via TCAV method}. The following provides a step-by-step workflow to (1) find the direction of the concept (2) evaluate these directions:
\begin{enumerate}
    \item Specify a concept, $\mathcal{C}$, e.g. life-sequence contains at least one \texttt{E16} diagnoses over year 2015,
    \item Randomly sample 10~000 life-sequences, $\mathbf{s}$, from the \textit{test} dataset: (1) for every life-sequence, $\mathbf{s}$, find a person-summary, $h(\mathbf{s}) = \mathbf{x}$ (where $h$ is the encoder part of the model), and  (2) calculate the gradient of output values, $\nabla f(\mathbf{x})$, with respect to the person-summary, $\mathbf{x}$, where $f$ is a decoder part of the model that takes a person-summary, $\mathbf{x}$, as an input and outputs logits,
    \item Randomly sample 3~000 life-sequences, $\mathbf{s}$, (from the \textit{validation} dataset) that satisfy the specifications of a concept $\mathcal{C}$ and calculate the person-summary, $\mathbf{x}$ for each sample (we will refer to this set as $\mathcal{D}_{\mathcal{C}}$
    \item Randomly sample 5~000 life-sequences (from the \textit{validation} dataset) that do not satisfy the specifications of a concept, $\mathcal{ C}$ and calculate the person-summary, $x$ for each sample (we refer to this set as $\mathcal{D}_{\neg \mathcal{C}}$
    \item Using stratified 5-fold cross-validation, find the optimal l2-regularisation parameter for the logistic regression on $\mathcal{D}_{\mathcal{C}} \cup \mathcal{D}_{\neg \mathcal{C}}$ datasets. The task for the logistic regression is to predict whether the life-sequence satisfies the concept  $\mathcal{C}$.
    \item Train 1~000 logistic regressions (with the optimal l2-regularisation parameter found in previous step) on the bootstrapped od $\mathcal{D}_{\mathcal{C}} \cup \mathcal{D}_{\neg \mathcal{C}}$ datasets,
    \item Find the orthonormal vector to each separating hyperplane found by the logistic regression. These normals are concept activation directions,  $h_C$.
    \item For each gradient vector,  $\nabla f(\mathbf{x})$, (step 1) find a dot product between every concept activation direction,  $h_C$ (step 6). The average of these values is a sensitivity score of a model to a particular concept. 
\end{enumerate}

As a baseline, we specify a random concept (i.e., no specifications). Then, during steps 2 and 3, we randomly sample sequences. Then, we use the Mann-Whitney U test \cite{mann1947test} to compare the distribution of the scores of baseline and distribution of the scores of a particular concept, $\mathcal{C}$.

\newpage
\subsection{Attention Mechanism} \label{app:attention_mechanism}
\begin{figure}[h]
    \centering
    \label{fig:attention_viz}
    \makebox[\textwidth][c]{\includegraphics[width=1\textwidth]{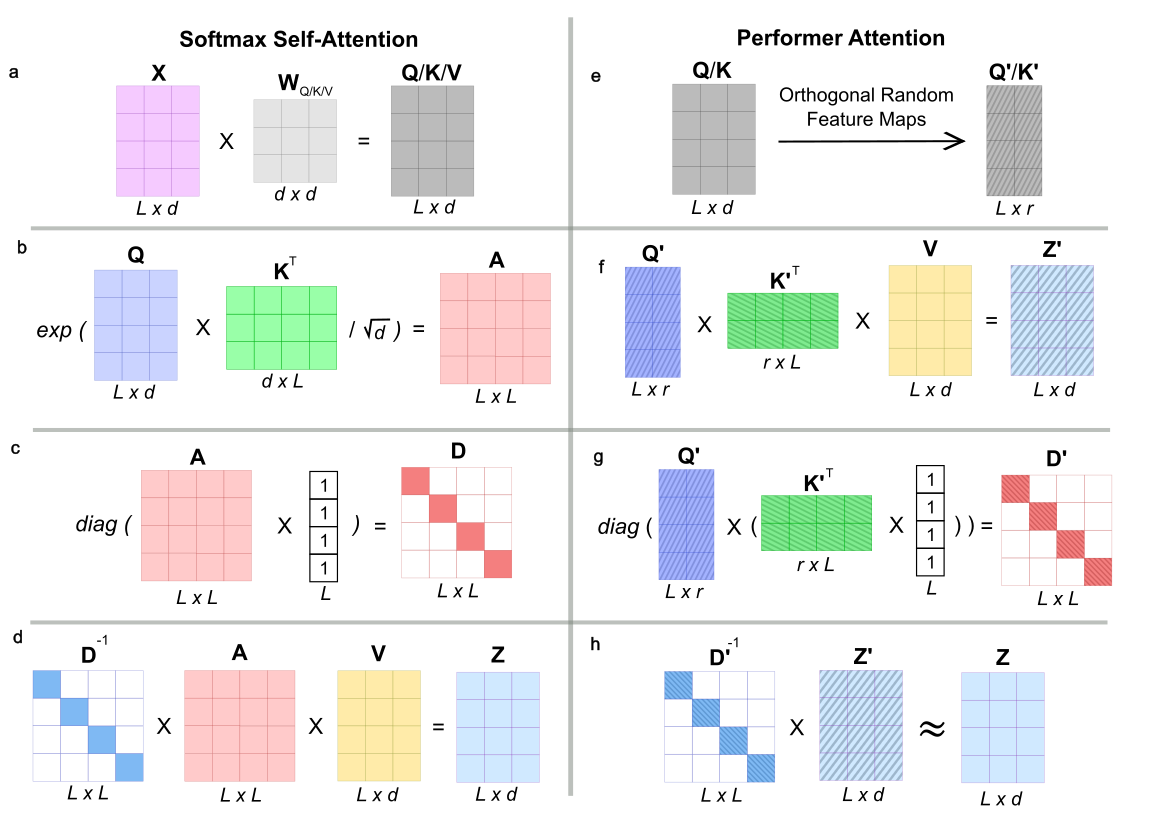}}
    \caption{Visualisation of the Attention Mechanism~\cite{vaswani2017attention}.~(\textbf{a}-\textbf{d}) Steps to compute the Softmax Self-Attention. (\textbf{a} and \textbf{e}-\textbf{h}) Steps to compute the FAVOR+ Attention (i.e., Performer-style Attention)~\cite{choromanski2020rethinking_performer} that approximates the Softmax Self-Attention output, the visualization omits the details for computing the Orthogonal Random Feature maps (for more details refer to~\cite{choromanski2020rethinking_performer}). You can find corresponding equations in Sec.~\ref{sec:methods}.}
\end{figure}

\subsection{Loss Functions} \label{app:loss_functions}
Based on the task, we use various loss functions to optimize the model.

\textbf{Pre-training}. Here we have two tasks -- Masked Language Model (MLM) and Sequence Order Prediction (SOP). For the MLM, we use \textit{cross-entropy loss}
\begin{equation}
\label{eq:ce_loss}
\mathcal{L}_{CE}(\mathbf{y}, \mathbf{x}) = -\sum y_i \log(f(\mathbf{x})_i)
\end{equation}

Here $y$ is a one-hot encoded true target, and $f(x)$ is the output of the model. For the SOP task, we use weighted \textit{cross-entropy loss with the label smoothing} \cite{muller2019does}

\begin{equation}
\label{eq:ce_loss_smooth}
\mathcal{L}_{CE-LS}(\mathbf{y}, \mathbf{x}) = (1-\alpha) \times -\sum w_i\,y_i\,\log(f(\mathbf{x})_i) + \frac{\alpha}{n} \times \sum w_i \,\log(f(\textbf{x}_i))
\end{equation}

Here $n$ is the number of classes, $w_i$ is a weight assigned to class $i$, and $\alpha$ is scalar to control for a mixture of CE and Label Smoothing components (i.e., $w=[1.1,10,10]$, $n=3$, and $\alpha = 0.1$). In Eq.~\ref{eq:ce_loss}, $\forall\, i\,w_i = 1$.

\textbf{Mortality Prediction and Emigration Prediction}. We use \textit{Asymmetric Cross-Entropy Loss} \cite{wang2021asymmetric}. It accounts for the fact that the unlabeled set might contain positive samples. It also does not require any knowledge of a prior for the frequency of positive samples in the unlabeled set.
\begin{equation}
\label{eq:asym_loss}
    \mathcal{L}_{ACE}(\mathbf{x}) = -\frac{1}{p} \sum_{i=1}^p \log \left(g(f\left(\mathbf{x}_i\right))_1\right)-\frac{1}{n-p} \sum_{i=p+1}^n \log \left(g(f\left(\mathbf{x}_i\right)+ \begin{bmatrix}
 c\\ 0
\end{bmatrix}^T)_0\right)
\end{equation}

In Eq.~\ref{eq:asym_loss}, $\mathbf{x}$ contains a batch of sequences,  $f$ is a model that outputs logits, $g(..)_1$ and $g(..)_0$ denote output (normalized scores) for positive and unlabeled classes (e.g. softmax function); while $n$ is a total number of samples in a batch and $p$ is the number of positive samples. $c \geq 0 $ is a constant added to a logit of a unlabeled sample; we choose it by optimising AUL metric.

\textbf{Predicting personality nuances}. We use mixture of \textit{Class Distance Weighted Cross-Entropy Loss} \cite{polat2022class}, \textit{Focal Loss} \cite{lin2017focal} and \textit{Label Smoothing} \cite{muller2019does}. 

\textit{Class Distance Weighted Cross-Entropy Loss}~\cite{polat2022class} handles imbalanced ordinal classification tasks. Instead of maximizing the likelihood of the 
true class, we minimize the likelihood of incorrect labels weighted by the absolute distance to the true class (Eq.~\ref{eq:cdw_loss}, where $y$ is a true label of a sample (\textit{not} one-hot encoded). CDW-CE has one hyperparameter -- $\alpha$, a distance penalty, which we set to 1.5.
\begin{equation}
\label{eq:cdw_loss}
\mathcal{L}_{CDW-CE}(y, \textbf{x}) = - \sum_{i=0}^{N-1}\textrm{log}(1 - f(\textbf{x})_i ) \times\left | i - y  \right |^{\alpha}
\end{equation}

\textbf{Focal Loss}~\cite{lin2017focal} is another version of cross-entropy loss that focuses on the samples that are hard to predict (Eq.~\ref{eq:focal_loss}), we set $\gamma = 5$. \textit{Label Smoothing}~(LS as in Eq.~\cite{szegedy2016rethinking}) penalizes \textit{overconfident} values (here $\mathbf{y}$ is a one-hot encoded target)

\begin{equation}
\label{eq:focal_loss}
\mathcal{L}_{F}(\mathbf{y}, \mathbf{x}) = - \sum \textrm{log}(\mathbf{y}_i - f(\mathbf{x})_i )^{\gamma}\,\textrm{log}(f(\mathbf{x})_i)
\end{equation}

The final loss function is presented in Eq.~\ref{eq:full_loss}. Without these modifications to the training procedure, the \texttt{life2vec} model converges to majority prediction. To train the model, we calculate $\mathcal{L}$ for each statement and then use the average of these losses to update the model.
\begin{equation}
\label{eq:full_loss}
\mathcal{L}(y, \mathbf{x}) = 0.3 \times \mathcal{L}_{CDW-CE}(y, \mathbf{x}) + \mathcal{L}_{F}(\texttt{onehot}(y), \mathbf{x}) +  0.1 \times \mathrm{LS}(\texttt{onehot}(y), \mathbf{x})
\end{equation}

\newpage
\subsection{Resampling based on the Instance Difficulty} \label{app:resampling}
We use the resampling strategy \cite{yu2022re} as one of the methods to account for a large imbalance of classes (Personality nuances task). This particular method is described for the case when the sample has only one target. In our case, we have four (i.e., four items). Thus, we had to adapt the method. The difficulty of an $i$th sample after $t$ steps is defined as \cite{yu2022re} 
$$
D_{i,T} = c + \sum\;d_{i,t}
$$
Given an $i$th sample, we calculate the difficulty $d_{i,t}^j$ according to Eq.4, 6, 7 in \cite{yu2022re} (where $t$ is a current epoch, and $j$ is a $j$th item). In several cases, it might happen that the value of $d_{i,t}^j$ is extremely high - even if the predictions for this sample are good on the subsequent steps, the $D_{i,T}$ is still going to be large. Thus, we set a threshold for the value of $d_{i, t}$, which equals 100. The  difficulty of the $i$th sample at the time step $t$ is 
$$
d_{i, t} = \min(\max\{d_{i,t}^1, \ldots, d_{i,t}^j\}, 100)
$$

After calculating all the difficulties at a step $t$, we apply robust scaling, where $Q$s are quantiles of difficulty scores at a step $t$
$$
\mathrm{RobustScaling}(d_{i,t}) = \frac{d_{i,t}}{Q_{0.75}-Q_{0.25}}
$$

Lastly, we change the calculation of $D_{i,T}$ by introducing the Exponential Weighted Average ($\alpha = .5$ and $D_{i,0} = c$) 

$$
D_{i,T} = \alpha \cdot d_{i,t} + (1-\alpha) \cdot D_{i, T-1}
$$

These operations help to stabilize the sampling weights (for our multi-target case). 

\newpage
\subsection{From Tabular Records to Life-Sequences}
\begin{figure}[h]
    \centering
    \makebox[\textwidth][c]{\includegraphics[width=1\textwidth]{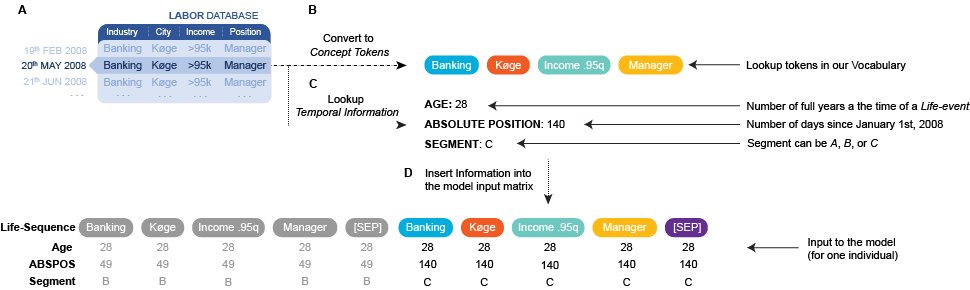}}
    \caption{The transformation of data from the Tabular format to the \texttt{life2vec} suitable format. (\textbf{A}) We start by looking up the next (chronologically) life-event in a person's history. (\textbf{B}) We convert relevant features to concept tokens of our vocabulary. (\textbf{C}) We lookup relevant positional information such as age, absolute temporal position, and segment of the life-event. (\textbf{D}) The information from the previous} 
\end{figure}

\newpage
\section{Visualisation of Embedding Spaces} \label{appx:embeddings}
\begin{figure}[ht!]
    \centering
    \label{fig:embeddings_all}
    \includegraphics[width=0.8\textwidth]{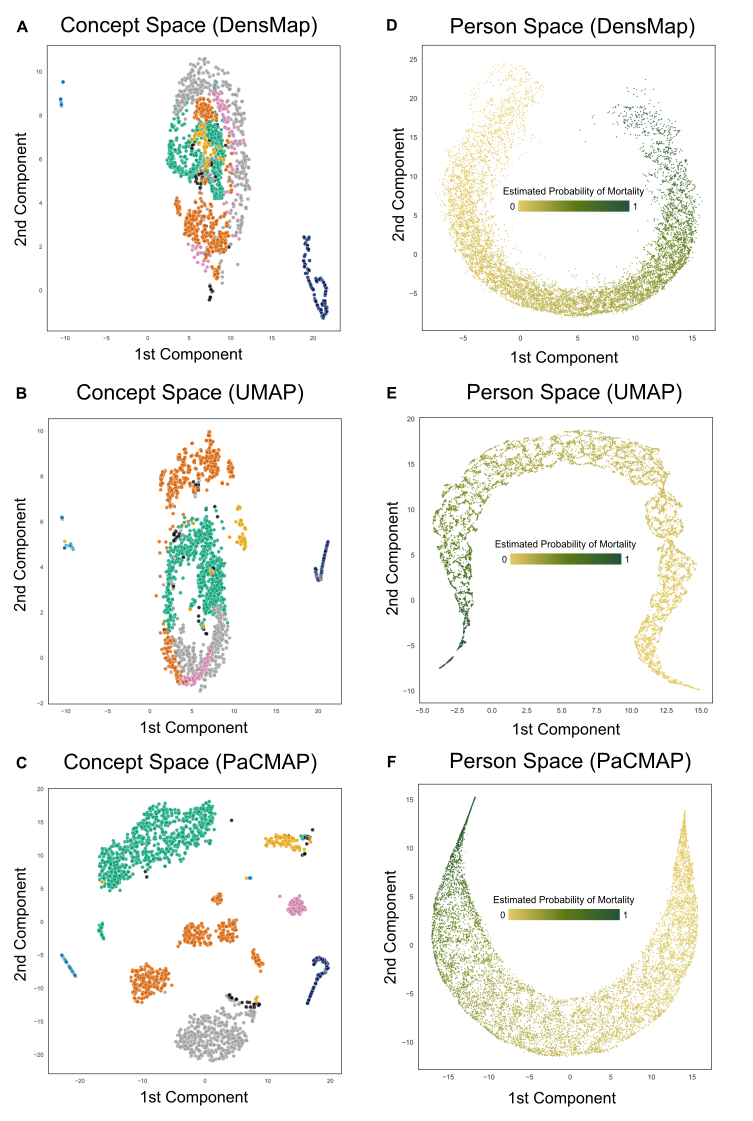}
    \caption{Projections of Concept Embedding Space (\textbf{A}-\textbf{C}) and Person Embedding Space (\textbf{D}-\textbf{F}). We visualize each space with three different projection methods: DenseMap, UMAP, and PaCMAP. In (\textbf{A}-\textbf{C}), concept tokens are colored based on the category.}
\end{figure}

\newpage
\section{Model Card: life2vec for Mortality Prediction}

The model card refers to \textbf{two} models:
\begin{enumerate}[nosep]
  \setlength\itemsep{0.05em}

    \item \texttt{life2vec} (\textit{base}) model pre-trained on the unsupervised masked language modeling and sequence order tasks, and
    \item \texttt{life2vec} (\textit{mortality}) model finetuned on the Mortality Prediction Task.
\end{enumerate}

\subsection{Model Details}
\textbf{Person or organization developing model}. \texttt{life2vec} (\textit{base} and \textit{mortality}) model is developed as part of the Nation-Scale Social Network Project. Specifically as part of the Ph.D. studies of Germans Savcisens at the Technical University of Denmark. The involved organizations are the Technical University of Denmark (Department of Applied Mathematics and Computer Science), Danmarks Statistik (Data Science Lab), and the University of Copenhagen (Copenhagen Center for Social Data Science). 

\textbf{Development period}. September 2020 - April 2023

\textbf{Model version}. \texttt{v1} (\textit{base}) and \texttt{v1} (\textit{mortality}).

\textbf{Model type}. Transformer-based deep neural network.

\subsubsection{Methods}

Architecture details of the \textit{base} model:
\begin{enumerate}[-,nosep]
    \setlength\itemsep{0.05em}
    \item \textbf{Architecture}: BERT-like encoder network
    \item \textbf{Optimization tasks}: Masked Language Modeling (MLM) and Sequence Order Classification (SOC). 
    \item \textbf{MLM Decoder Network}: Two-layer neural network (pools representation of masked tokens). The final layer is tied to the Concept token embedding matrix ( SI~\ref{app:architecture}).
    \item \textbf{SOC Decoder Network}: Two-layer neural network (pools \texttt{SEP} representation, SI~\ref{app:architecture}).
    \item \textbf{Attention mechanism}: Performer-style attention heads, with local softmax-attention heads (SI~\ref{app:attention_mechanism}).
    \item Other \textbf{architecture modifications} include ReZero \& ScaleNorm normalization, Swish, Input-Output Embedding Tying, \texttt{time2vec}-based encoding of temporal information, sequence order prediction task, and cross-entropy loss with the label smoothing. These modifications ensure fast convergence and relatively small size of the model.
    \item \textbf{Optimization strategy}: AdamW Optimizer with the OneCycle LR Annealing. Trained for 30 epochs, where one epoch covers 30~000 randomly sampled (and augmented) sequences from the training dataset.
\end{enumerate}

Architecture details of the \textit{mortality} model:
\begin{itemize}[-,nosep]
  \setlength\itemsep{0.05em}
    \item \textbf{Architecture}: BERT-like encoder network (aka pre-trained \textit{base} model)
    \item \textbf{Optimization task}: Binary mortality prediction task (early mortality within the next four years).
    \item \textbf{Classification decoder}: Two-layer network with the weighted averaging of concept representations (SI~\ref{app:architecture}).

    \item Other \textbf{architecture modifications}: Asymmetric Cross-Entropy (SI:~\ref{app:loss_functions}) for the Positive Unlabeled data sets and Sigsoftmax (as an alternative to Softmax). 
    \item \textbf{Optimisation strategy}: RAdam optimizer with exponential LR annealing ($\gamma = 0.8$). Base LR for the decoder is 0.01, and LR for each consecutive encoder layer reduces by 5\%. Token embeddings are frozen except for the \texttt{[CLS], [SEP], [UNK]} tokens.  Temporal and Segment Embeddings are not frozen. For the decoder network, we set weight decay to 0.01. For the encoder layers, we set weight decay to 0.001. 
    \item \textbf{Data Sampling}: We re-sample positive and negative samples to get approximately an equal fraction of both targets. 
\end{itemize}

\textbf{Fairness Constraints}. None.

\textbf{License}: Not for public use or distribution.

\textbf{Primary intended uses}. The following information covers both \textbf{base} and \textbf{mortality} models:
\begin{itemize}[-,nosep]
  \setlength\itemsep{0.05em}
    \item Use for scientific and research purposes only,
    \item Verify the validity of \texttt{NLP} inspired \textit{Socio-economic} data representation,
    \item Verify the validity and performance of transformer-based architectures in the context of longitudinal socio-economic data
    \item Explore interactions between life-events and outcomes (i.e., mortality prediction) \textit{on a global scale}. 
    \item Use person-summaries as node-features in the Large-scale \textit{Danish} population graphs
    \item Use person-summaries embeddings to study causal relationships between life-events.
\end{itemize}

\textbf{Primary intended users}. The following information covers both \textbf{base} and \textbf{mortality} models:
\begin{enumerate}
  \setlength\itemsep{0.1em}
    \item Denmark-based academics in Computational Social Science, Economics, Healthcare, Sociology, or Network Science.
    \item Employees of Denmark Statistics (specifically Data Science Lab).
\end{enumerate}

\textbf{Out-of-scope use cases}. The following information covers both \textbf{base} and \textbf{mortality} models:
\begin{enumerate}
    \setlength\itemsep{0.1em}
    \item \textbf{Not intended} as a tool to make judgments about specific individuals.
    \item \textbf{Not intended} for a public release or deployment in governmental or private institutions.
\end{enumerate}

\subsection{Factors}
This section describes the factors (e.g., groups, socio-economic attributes, sequence structure, etc.) that might lead to discrepancies in the model performance. \textbf{The section covers the \textit{mortality} model.}

Potential \textbf{relevant factors} are
\begin{enumerate}[nosep]
    \setlength\itemsep{0.1em}
    \item \textbf{Level of interaction with the healthcare system} -- the fact that people use and consult healthcare providers with different frequencies (e.g., a person avoids interaction with the healthcare system or interacts only in severe cases),
    \item \textbf{Socio-demographic attributes}: age, sex, and residency status (different groups, e.g., immigrants, ex-pats, natural-born citizens, and other residents, might have various access to public services and various sets of opportunities and limitations),
    \item \textbf{Sequence Length} -- longer sequences might contain more information (that model can use).
    \item \textbf{Data drift and time} -- we cannot guarantee the robust performance of the model beyond 2020 (e.g., COVID and human behavior). 
    \item \textbf{Cause of death}
\end{enumerate}

\textbf{Evaluation factors}. Since \textbf{age} and \textbf{sex} is highly correlated with mortality outcomes, we want to evaluate the model's performance on unitary and intersectional splits of these groups (to probe the \texttt{life2vec}'s sensitivity to these features). Regarding the \textbf{residency status}, we are limited to the split based on the birthplace (i.e., in Denmark or outside). Thirdly, we want to evaluate how robust \texttt{life2vec} is regarding various life-sequences structures (aka length- and event-wise). Thus, we look at the \textbf{number of health-related events} in a sequence and the \textbf{length of the sequence}. Lastly, we do not have access to data beyond 2020. Thus, we cannot estimate the effect of the data drift on the \texttt{life2vec} model. However, we can evaluate how well the model predicts \textit{distant} deaths. 

\subsection{Metric}
\textbf{Pre-training}. We look at the perplexity score to evaluate and choose the most optimal \texttt{life2vec} (base) model (not presented in the model card). 

\textbf{Mortality Prediction}. We frame the mortality prediction task as a positive-unlabeled problem. To optimize the \texttt{life2vec} (mortality) model, we use Area-Under-the-Lift (AUL), i.e., the early-stopping mechanism uses the AUL score.  The primary performance evaluation metric for the \texttt{life2vec} model is Corrected Mathew's Correlation Coefficient (C-MCC) with a 95\% Confidence Interval (we use bootstrapping). We use correction to account for the unlabeled samples in the test dataset. Along, we provide the corrected Balanced Accuracy and Corrected F1-Score (refer to Tab.~\ref{tab:eos_metric}). All metric is reported at the .5 probability cutoff (not applicable to the AUL).

All the metrics presented in this model card are based on the test subset.

\begin{table}[h]
\centering
\caption{Corrected Matthew's Correlation Coefficient (C-MCC) and Area under the Lift (AUL) on the Mortality Prediction Task (comparison of different baseline models).  95\%-Confidence intervals for the MCC based on the stratified bootstrapping. In both cases, the higher value is preferred.}
\label{tab:eos_metric}
\small
\begin{tabular}{@{}lllllc@{}}
\toprule
\multicolumn{1}{c}{\textbf{Model}} &
  \multicolumn{1}{c}{\textbf{MCC, 95\%-CI}} &
  \multicolumn{1}{c}{\textbf{AUL}} &
  \multicolumn{1}{c}{\textbf{Accuracy, 95\%-CI}} &
  \multicolumn{1}{c}{\textbf{F1-Score, 95\%-CI}} &
  \multicolumn{1}{c}{\textbf{Model Size}} \\ \midrule
\textbf{L2V}   & \textbf{0.413 {[}0.410, 0.422{]}} & \textbf{0.845} & 0.788 {[}0.782, 0.794{]} & 0.443 {[}0.435, 0.451{]} & 8.4m\\
RNN-GRU        & 0.369 {[}0.361, 0.378{]}          & 0.834          & 0.778 {[}0.771, 0.783{]} & 0.395 {[}0.389, 0.402{]} & 1.5m\\
FFNN           & 0.340 {[}0.332, 0.348{]}          & 0.822          & 0.768 {[}0.762, 0.774{]} & 0.345 {[}0.339, 0.350{]} & 8.4m\\
Logistic Reg   & 0.149 {[}0.142, 0.155{]}          & 0.735          & 0.639 {[}0.633, 0.645{]} & 0.201 {[}0.198, 0.204{]} & 2.0k\\
Life Tables    & 0.059 {[}0.051, 0.066{]}          & 0.650          & 0.555 {[}0.548, 0.562{]} & 0.161 {[}0.158, 0.164{]} &  3\\
Random         & -0.005 {[}-0.011, 0.002{]}        & 0.497          & 0.496 {[}0.489, 0.503{]} & 0.132 {[}0.128, 0.135{]} & - \\
Majority Class & 0.0                               & 0.497          & 0.5                      & -                        & - \\ \bottomrule
\end{tabular}
\end{table}

\textbf{Quantitative Analysis}. We estimate the C-MCC score on the test data split (not the full one, but a random subsample of 20~000 people). See Fig. 6-9.

\begin{figure}[h]
    \centering
    \makebox[\textwidth][c]{\includegraphics[width=0.4\textwidth]{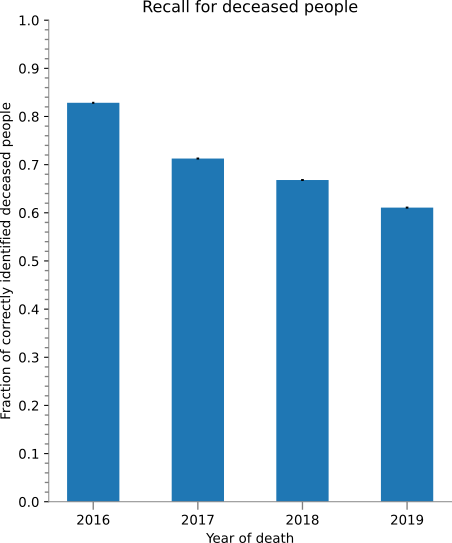}}
    \caption{The \texttt{life2vec}'s recall is based on the period between the day of prediction and day of death. The performance degrades as we get further away from the 31st of December 2015.}
\end{figure}

\begin{figure}[h]
    \centering
    \makebox[\textwidth][c]{\includegraphics[width=\textwidth]{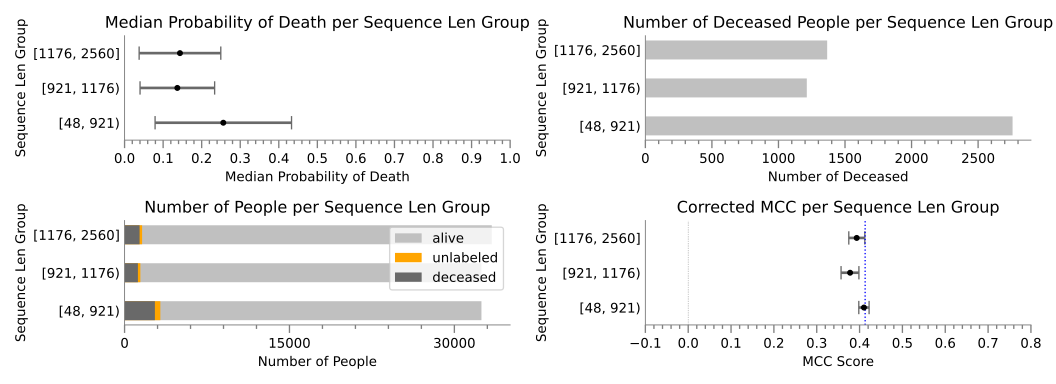}}
    \caption{Detailed evaluation of the \texttt{life2vec} model based on the \textbf{sequence length}. The length of the sequence does not seem to impact the performance of the model.}
\end{figure}

\begin{figure}[h]
    \centering
    \makebox[\textwidth][c]{\includegraphics[width=1\textwidth]{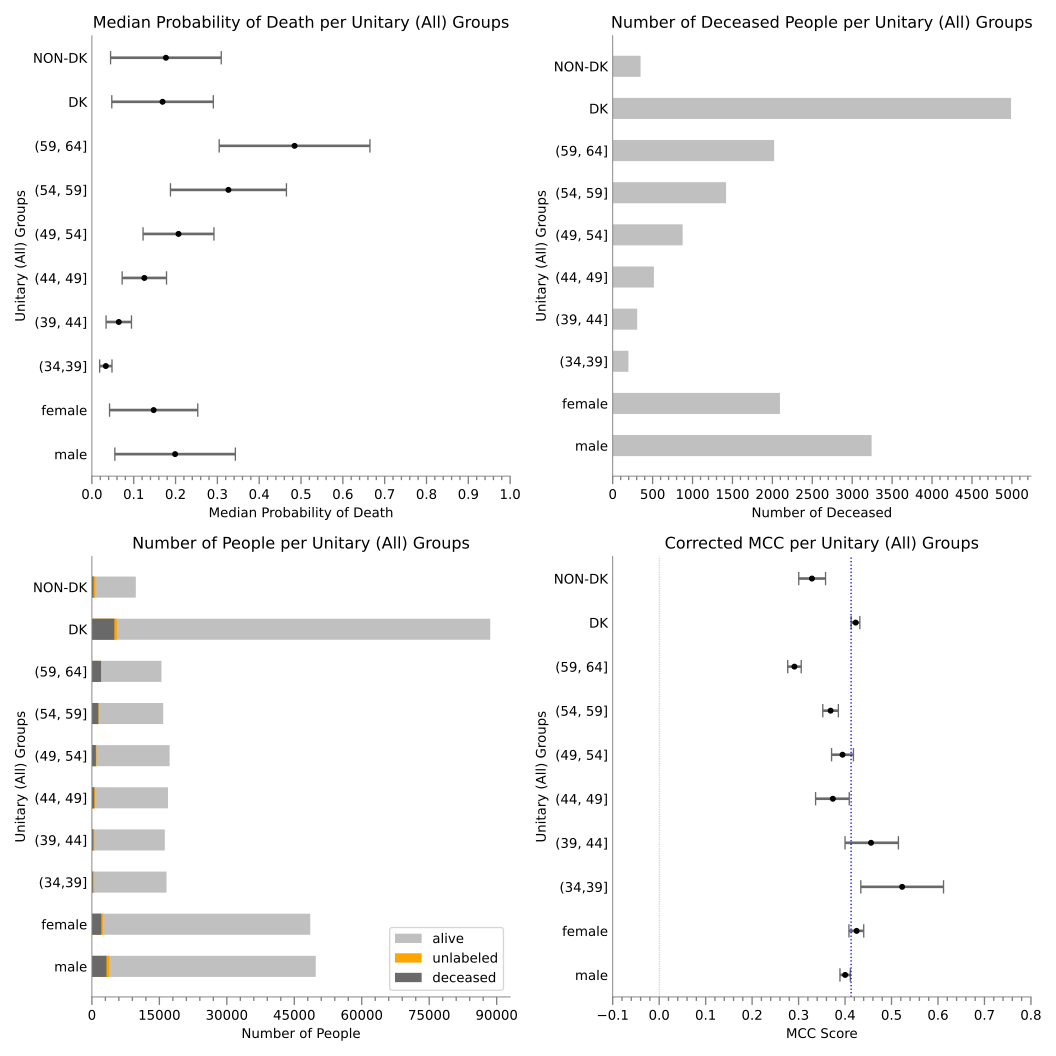}}
    \caption{Detailed evaluation of the \texttt{life2vec} model based on the \textbf{socio-demographic} attributes. The sequence length does not seem to impact the model's performance. \textbf{Age} -- generally, older people have a higher probability of death. At the same time, the performance metric is worse for older people. \textbf{Sex} -- the model's performance is similar regarding sex attributes. \textbf{Residency} -- we can see a large difference between DK and NoN-DK groups, which might be connected to the imbalanced representation of groups. }
\end{figure}

\begin{figure}[h]
    \centering
    \makebox[\textwidth][c]{\includegraphics[width=1\textwidth]{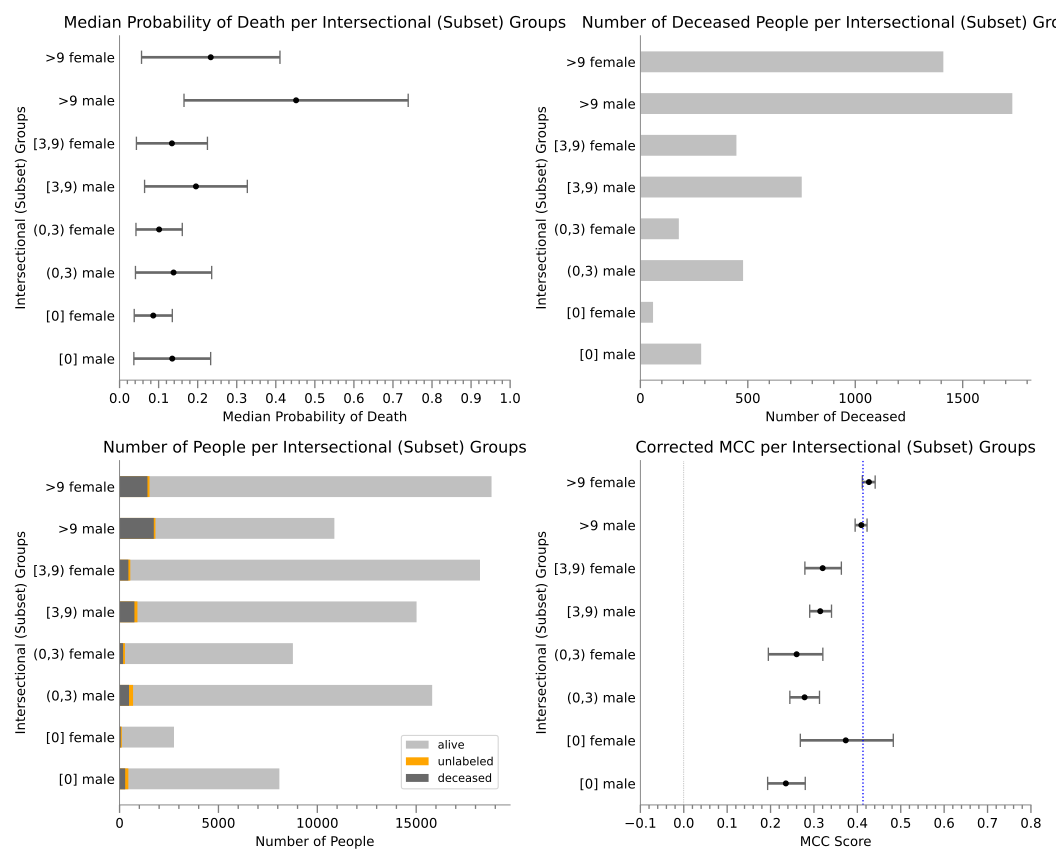}}
    \caption{Detailed evaluation of the \texttt{life2vec} model based on the intersection of \textbf{sex} and the \textbf{number of health events}. The results confirm that the level of interaction with the healthcare system \textit{does have} an impact on the quality of predictions. }
\end{figure}

\subsection{Data}
Labor \cite{amrun_data} and health data \cite{lynge2011danish, world1992icd} are provided by Danmarks Statistik (DST). These datasets include socio-economic, longitudinal information about the residents of Denmark. The use of the data \cite{amrun_data,lynge2011danish, world1992icd} is regulated by (1) the EU Regulation on European Statistics, (2) the General Data Protection Regulation (GDPR), (3) the Danish Data Protection Act, (4) the Danish Public Administration Act, (5) the Danish Access to Public Administration Files Act, (6) the Danish Criminal Code, and (7) the Act on Statistics Denmark (DST). DST ensures that data of Danish residents and businesses is used \textbf{only for scientific purposes}.

\textbf{Preprocessing}. Refer to the Methods Section in the Original Paper.

\textbf{Data split}. We split data into training, validation, and test datasets (completely at random). \textbf{Training} subset is used to optimize the model. \textbf{Vaidation} subset is used to evaluate the model's performance at a specific epoch - we terminate the model's training if the performance metric on the validation data does not improve. \textbf{Test} subset estimates the final model performance.

\textbf{Ethical Considerations}. Refer to the Methods Section in the Original Paper.

\end{document}